\pdfoutput=1

\documentclass[11pt]{article}

\usepackage[table,xcdraw]{xcolor}

\usepackage{acl}

\usepackage{times}
\usepackage{latexsym}

\usepackage{amsfonts}
\usepackage{amsmath}
\usepackage{graphicx,subfigure}
\usepackage{booktabs}
\usepackage{multirow}
\usepackage{url}
\usepackage{tabularx}
\usepackage{makecell}
\usepackage{multirow}
\usepackage{graphicx}
\usepackage{comment}

\newcolumntype{h}{X}
\newcolumntype{s}{>{\hsize=.62\hsize}X}
\newcolumntype{t}{>{\hsize=.64\hsize}X}

\usepackage[T1]{fontenc}

\usepackage[utf8]{inputenc}

\usepackage{microtype}

\usepackage{inconsolata}

\title{FineSurE: Fine-grained Summarization Evaluation using LLMs}

\author{Hwanjun Song$^{1}$\thanks{~~Corresponding Author.},~ Hang Su$^{2,}$\thanks{~~This work conducted independently and is not related to the author(s) position at Amazon.}~,~ Igor Shalyminov$^{2,\dagger}$,~ Jason Cai$^{2,\dagger}$,~ Saab Mansour$^{2,\dagger}$\\
$^{1}$Korea Advanced Institute of Science and Technology\\ $^{2}$AWS AI Labs\\
songhwanjun@kaist.ac.kr\\ \{shawnsu, shalymin, cjinglun, saabm\}@amazon.com}

\newcommand{\algname}{{FineSurE}} 
\newcolumntype{L}[1]{>{\raggedright\let\newline\\\arraybackslash\hspace{0pt}}m{#1}}
\newcolumntype{X}[1]{>{\centering\let\newline\\\arraybackslash\hspace{0pt}}p{#1}}
\newcolumntype{Y}[1]{>{\raggedleft\let\newline\\\arraybackslash\hspace{0pt}}m{#1}}

\begin{document}
\maketitle
\begin{abstract}


Automated evaluation is crucial for streamlining text summarization benchmarking and model development, given the costly and time-consuming nature of human evaluation. Traditional methods like ROUGE do not correlate well with human judgment, while recently proposed LLM-based metrics provide only summary-level assessment using Likert-scale scores. This limits deeper model analysis, e.g., we can only assign one hallucination score at the summary level, while at the sentence level, we can count sentences containing hallucinations.
To remedy those limitations, we propose \algname{}, a fine-grained evaluator specifically tailored for the summarization task using large language models (LLMs). It also employs completeness and conciseness criteria, in addition to faithfulness, enabling multi-dimensional assessment. We compare various open-source and proprietary LLMs as backbones for \algname{}. In addition, we conduct extensive benchmarking of \algname{} against SOTA methods including NLI-, QA-, and LLM-based methods, showing improved performance especially on the completeness and conciseness dimensions. The code is available at \url{https://github.com/DISL-Lab/FineSurE-ACL24}.

\end{abstract}

\section{Introduction}

Text summarization stands out as an important task in natural language processing, aiming to generate a condensed summary of a provided text while retaining its essential information\,\cite{gupta2019abstractive,song2023enhancing}. 
Despite the enhanced quality of summaries produced by LLMs, the development of automated methods for evaluation remains a challenge \cite{kryscinski2020evaluating, maynez2020faithfulness}. 
Conventional \emph{reference-based} metrics, such as ROUGE\,\cite{lin2004rouge}, have exhibited a \emph{weak} correlation with actual human judgments\,\cite{liu2023gpteval}. Consequently, human evaluation remains an essential step for accurately assessing the quality of generated summaries, even considering its inherent costs and time-consuming nature. 

Recently, the need for better automatic evaluators has become an important research topic, aiming to streamline evaluation processes and ease manual efforts in model development \cite{gao2023human}. This effort provides valuable insights into {whether generated summaries align with predefined quality standards}, including aspects like faithfulness. 
Various approaches have been explored, including approaches based on {neural language inference\,(NLI)}\,\cite{laban2022summac} and {question-answering\,(QA)}\,\cite{fabbri2022qafacteval, zhong2022towards}. In addition, LLMs have recently proven their potential to be an \emph{automated} tool for human-like evaluation\,\cite{liu2023gpteval, wang2023chatgpt}. The latest LLM-based method, G-Eval\,\cite{liu2023gpteval}, demonstrated a Spearman correlation coefficient of {over 0.5} with Likert-scale human judgments on the news domain using GPT-4.

\begin{figure*}[t!]
\begin{center}
\includegraphics[width=16.1cm]{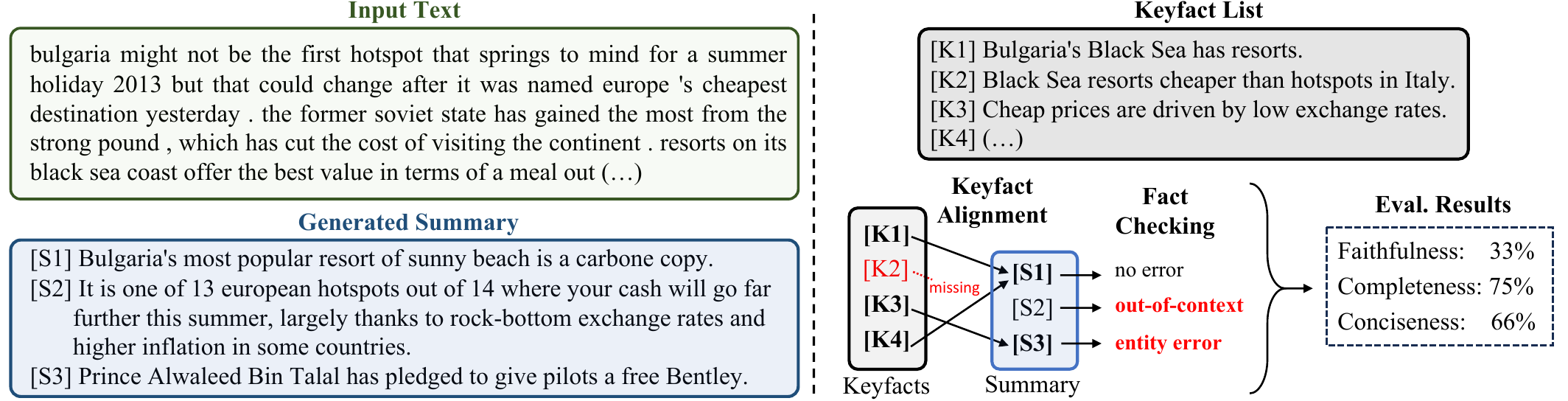}
\end{center}
\vspace*{-0.3cm}
\caption{\textbf{\algname{} framework:} the given summary is evaluated by conducting the two tasks of fact checking and keyfact alignment. In this specific example, the faithfulness score is 33\%, since only one out of the three summary sentences is factually correct; the completeness score is 75\%, since three out of the four keyfacts align with the summary; and the conciseness score is 66\%, since two out of the three sentences are related to the keyfacts.}
\vspace*{-0.35cm}
\label{fig:framework}
\end{figure*}

Despite these advancements, we contend that the current LLM-based automated methods still fall short in achieving precise evaluation, primarily attributed to the \emph{coarse-grained} evaluation pipeline and the \emph{ambiguity} in evaluation dimensions. 
Specifically for coarse-grained evaluation, the evaluation dimensions--namely faithfulness, coherence, and relevance~\footnote{We omit fluency assessment as modern AI models typically generate highly fluent outputs\,\cite{liu2023gpteval}.}--are frequently assessed at the summary-level, resulting in {Likert-scale} scores for each summary\,\cite{gao2022dialsummeval, shen2023opinsummeval, liu2023gpteval, wang2023chatgpt}. This Likert-scale scoring method lacks fine-grained information on errors in generated summaries. For instance, it does not provide a breakdown of the number of summary sentences with quality issues or specify the types of mistakes present in each sentence.
Furthermore, regarding ambiguity, the evaluation of coherence and relevance is hindered by the lack of clarity in their definition of "the collective quality of all sentences" and "the selection of important content from the source"\,\cite{fabbri2021summeval, shen2023large}. 
Given that human can encounter challenges in evaluating summaries, it is inappropriate to expect a neural model to provide accurate and objective assessments. 
Hence, there is a need to develop a more precise evaluation framework that results in a more detailed assessment with clearly defined evaluation dimensions.

In this paper, {we present \textbf{\algname{}} (\textbf{Fine}-grained \textbf{Su}mma\textbf{r}ization \textbf{E}valuation) using LLMs}, a novel automated approach designed to evaluate the summarization quality at a \emph{fine-grained} level based on summary sentences or keyfacts\footnote{A keyfact refers to a concise sentence conveying a single key information, comprising at most 2-3 entities, also referred to as a semantic content unit\,\cite{bhandari2020re}. The keyfact list can be generated automatically or by humans.}, as depicted in Figure \ref{fig:framework}. 
We aim to evaluate summaries using this framework along three vital criteria: the \emph{faithfulness} of minimizing factuality errors, the \emph{completeness} of encompassing the majority of keyfacts, and the \emph{conciseness} of avoiding unnecessary details. Thus, our framework entails executing two finely grained procedures utilizing LLMs: \textbf{(1) fact checking} involves identifying specific factuality errors present in each summary sentence and \textbf{(2) keyfact alignment} focuses on aligning each keyfact with all summary sentences from which they are inferred. 
We leverage the outcomes from both procedures to calculate \emph{precise percentage} scores, offering a more detailed assessment than Likert-scale scoring. 
This fine-grained approach enables us to analyze the quality issues of generated texts at both the sentence and keyfact-level. 

On top of keyfact alignment, two dimensions (completeness and conciseness) can serve as better replacements for coherence and relevance\,\cite{fabbri2021summeval}, as they evaluate two key aspects of a good summary, assessing the comprehensive inclusion and density of key information while excluding irrelevant content.

To summarize, our main contributions are as follows: (1) we argue that LLM-based summarization suffers from hallucination, information emission and verbosity hence requiring revisiting the evaluation dimensions, (2) we suggest three metrics targeting LLM output characteristics and tackling the aforementioned problems including faithfulness, completeness and conciseness, (3) we propose a novel automated evaluation framework - \algname{}, based on keyfact lists and using LLMs to generate the keyfacts, align them to the summary sentences and categorize the errors automatically, (4) we compare various open-source and proprietary LLMs to power \algname{} and analyze their correlation to human judgment at the summary and system levels, (5) we provide comprehensive results comparing our method with similarity-based, NLI-based, QA-based, and LLM-based automated methods and show improved human correlation for \algname{} over state-of-the-art methods.


\section{Related Work}\label{sec:rel}

Efforts to assess the quality of texts generated by language models have led to numerous initiatives in designing effective automated evaluators across various research directions\,\cite{lin2004rouge, fabbri2022qafacteval, zhong2022towards, wang2023chatgpt}. 

\paragraph{Similarity-based Evaluator.} The evaluation of generated text can be measured by the n-gram overlap with the reference text, employing metrics such as ROUGE \cite{lin2004rouge}, BLEU \cite{papineni2002bleu}, and METEOR \cite{banerjee2005meteor}. In contrast to relying on exact matches, several evaluators have leveraged token similarity through contextual embeddings like BERTScore \cite{zhang2019bertscore}, MoverScore \cite{zhao2019moverscore}, and BARTScore \cite{yuan2021bartscore}. 
However, these evaluators lack human correlation and a multi-dimensional assessment of text quality akin to real human evaluation, as they typically produce a single-dimensional score based on text similarity.

To assess text quality, primarily focusing on checking factual consistency, task-specific evaluators utilizing NLI and QA have been explored.

\vspace*{-0.1cm}
\paragraph{NLI-based Evaluator.} This task involves fact-checking and verification by retrieving relevant evidence from the input text to support the claim made in the generated text \cite{glover2022revisiting,honovich2022true, utama2022falsesum}. DAE\,\cite{goyal2020evaluating} introduced the dependency arc entailment formulation, offering a more fine-grained approach to faithfulness evaluation.
SummaC \cite{laban2022summac} presented a lightweight model that facilitates NLI by segmenting input text into sentence units and aggregating scores between pairs of sentences. Despite their enhanced performance, they only focus on assessing faithfulness.

\vspace*{-0.1cm}
\paragraph{QA-based Evaluator.} 
This involves generating plausible questions from the reference text and then answering these questions considering the generated text\,\cite{scialom2021questeval, chen2021factuality}. QAGS \cite{wang2020asking} and QAFactEval \cite{fabbri2022qafacteval} enhanced the accuracy of faithfulness evaluation, surpassing other similarity- and NLI-based evaluators in text summarization. UniEval \cite{zhong2022towards} proposed a unified evaluator capable of assessing multi-dimensional evaluation of text generation through the QA task. In text summarization, it evaluates four aspects: faithfulness, coherence, relevance, and fluency. Generally, these methods require training a neural model to generate questions and their corresponding answers.

\vspace*{-0.05cm}
\paragraph{LLM-based Evaluator.} With the emergence of LLMs, there is a move toward utilizing them as reference-free automated evaluators in diverse scenarios\,\cite{shi2023llm, lin2023llm, chen2023exploring, fu2023gptscore}. 
Recently, a few efforts have been made to evaluate faithfulness using edited text\,\cite{laban2023summedits}, atomic facts\,\cite{min2023factscore}, and external knowledge base\,\cite{feng2023factkb}, as well as to assess multi-dimensional aspects\,\cite{liu2023gpteval, shen2023large, wang2023chatgpt}. TrueTeacher\,\cite{gekhman2023trueteacher} employs LLMs to generate large-scale synthetic data for fact verification.
Although LLMs have shown promise as evaluators, they currently lack a fine-grained assessment and primarily focus on addressing faithfulness, without considering other important dimensions for high-quality summaries.

Unlike prior studies, we define crucial aspects for a detailed evaluation using LLMs and introduce a new fine-grained evaluation framework called \algname{}. This framework addresses numerous open questions regarding the capabilities of LLMs, including sentence-level fact-checking, classification of error types, and keyfact-level alignment.

\section{\algname{} Framework}\label{sec:method}


\subsection{Evaluation Dimensions}


LLMs enhance the quality of summarization, but they rather suffer from hallucination, information emission and verbosity\,\cite{ji2023survey, saito2023verbosity}, requiring revisiting evaluation dimensions. 
Therefore, we advocate for a thorough assessment of the two evaluation criteria, "completeness" and "conciseness," in addition to "faithfulness."
These two dimensions can effectively assess both information emission and verbosity while also complementing each other in evaluating information inclusion and summary succinctness. 


\vspace*{-0.2cm}
\begin{itemize}
\setlength{\itemindent}{-1em}
\item \textbf{Faithfulness}: The summarizer does not manipulate the information in the input text (i.e., intrinsic) and add any information not directly inferable from the input text (i.e., extrinsic). \vspace*{-0.25cm}
\item \textbf{Completeness}: The summarizer ensures the inclusion of all keyfacts from the input text in the output summary.\vspace*{-0.25cm}
\item \textbf{Conciseness}: The summarizer refrains from incorporating information outside the keyfacts in the output, maintaining a succinct and focused summary.\vspace*{-0.2cm}
\end{itemize}

Note that, adhering to the precise definition of faithfulness in the recent work \cite{pagnoni2021understanding}, we categorize error types into a total of seven categories, with "out of context" as an extrinsic error, and "predicate," "entity," "circumstance," "coreference," "discourse link," and "grammatical" as intrinsic errors. See examples in Appendix \ref{sec:category_example}.

\subsection{Evaluation Pipeline}
We discuss the evaluation pipeline implementing the dimensions discussed previously.
We employ LLMs as a tool to conduct fact checking and keyfact alignment tasks. Specifically, we design two prompts tailored for the two tasks, as shown in Figures \ref{fig:best_prompt_factuality}-\ref{fig:best_prompt_keyfact} of Appendix \ref{sec:main_prompt}. All prompts are customized to generate outputs in JSON format, enhancing the success ratio of following our instructions and facilitating parsing. The detailed analysis of the success ratio is provided in Section \ref{sec:success_ratio}.

\paragraph{Task 1. Fact Checking.} Figure \ref{fig:best_prompt_factuality} illustrates our prompt and its expected JSON output for fact checking. We convert the problem of fact checking into a \emph{categorization problem} involving nine categories. These include the seven factuality errors, along with an additional category "other error" for errors outside the seven errors, and an additional category "no error" for cases where no error was detected. Therefore, given a pair of input text and model summary, the LLM is expected to output the \emph{error type} classified into one of the nine categories for each sentence along with a concise reason.

\paragraph{Task 2. Keyfact Alignment.} Figure \ref{fig:best_prompt_keyfact} shows our prompt and its expected JSON output for keyfact alignment. We address the alignment problem through \emph{keyfact matching}, a process that involves two sequential tasks: verifying if each keyfact is inferred from the summary and, if affirmative, specifying the line numbers for all the corresponding summary sentences. Thus, given a pair of keyfact list\footnotemark[3] and model summary, the output should be the binary label and the list of line numbers of all summary sentences matched for each keyfact.


\paragraph{Parsing and Scoring.} The evaluation scores are computed using the results from the two tasks. Given a document $D$, let $S =\{s_1, \dots s_N\}$ be the generated summary with $N$ sentences. By the fact checking task, we identify a subset of $S_{fact} \subseteq S$, which consists solely of summary sentences marked "no error". Then, the {percentage score} of faithfulness on $S$ is determined by:
\begin{equation}
{Faithfulness}(D, S) = |S_{fact}| / |S|. 
\label{eq:faithfulness_score}
\end{equation}

Let $K=\{k_1, \dots, k_M\}$ be the list of keyfacts with a size of $M$. Through the keyfact alignment, we construct a bipartite graph $M=(K, S, E)$, where the edge set $E = \{(k, s):  k \rightarrow s~|~k \in K \wedge s \in S \}$ and $k \rightarrow s$ indicates that the keyfact $k$ aligns with the summary sentence $s$. Then, the percentage scores of completeness and conciseness on $S$ are computed at the summary level by:
{
\begin{equation}
\begin{gathered}
\!\!\!\!{Completeness}(K, S)\! =\! \big|\{k|(k,s)\!\in E\}\big| / |K|\!\!\\
\!\!\!\!{Conciseness}(K, S) \!= \!\big|\{s|(k,s)\!\in E\}\big| / |S|,\!\!
\end{gathered}
\label{eq:other_scores}
\end{equation}
where the operator $|\cdot|$ returns the number of unique items within the provided set. Intuitively, the two scores represent completeness, indicating the degree to which keyfacts are included in the summary, and conciseness, reflecting the density of relevant sentences aligning with given keyfacts. 
Moreover, unlike existing LLM-based methods \cite{liu2023gpteval, wang2023chatgpt, shen2023large}, we provide more detailed information about the error type associated with each sentence and the alignment of each keyfact with summary sentences.

\footnotetext[3]{The list of keyfacts is provided by humans; if unavailable, it can be automatically derived from the reference summary. See Appendix \ref{sec:deatail_keyfact_extraction} for details.}

\subsection{Prompt Engineering}

We explore various prompt engineering strategies\,\cite{wei2022chain, yu2023towards} to identify the most suitable one for our evaluation pipeline:
\vspace*{-0.2cm}
\begin{itemize}
\setlength{\itemindent}{-1em}
\item \textbf{Basic Prompt}: A default question prompt in plain text, e.g., is the summary sentence supported by the transcript?
\vspace*{-0.25cm}
\item \textbf{Instruction}: The prompt is provided using a step-by-step instruction using "Instruction:".
\vspace*{-0.25cm}
\item \textbf{Categorization}: The prompt solves a categorization task by providing target categories.  
\vspace*{-0.75cm}
\item \textbf{Reasoning}: The prompt uses a chain-of-thought approach, incorporating a reasoning step.
\vspace*{-0.25cm}
\item \textbf{Evidence Mapping}: The prompt requests an exact quote from the input to confirm the decision made by LLMs.
\vspace*{-0.25cm}
\end{itemize}

Combining all the above techniques was not always superior. Evaluation prompts are recommended to use instruction format with categorization and reasoning for faithfulness evaluation, as in Figure \ref{fig:best_prompt_factuality}, and only instruction format for completeness and conciseness evaluation, as in Figure \ref{fig:best_prompt_keyfact}. See the detailed ablation in Appendix \ref{sec:prompt_engineering}. 

\begin{table*}[t]
\begin{center}
\footnotesize
\begin{tabular}{|L{1.5cm} |L{2.6cm} |X{2.25cm} |X{2.1cm} X{2.1cm} X{2.0cm}| }\toprule
Direction & Method & Sentence-level & \multicolumn{2}{c}{Summary-level} & System-level \\
 &  & bAcc\,($\uparrow$) & \!\!Pearson Corr\,($\uparrow$)\!\! & \!\!Spearman Corr\,($\uparrow$)\!\! & Rank Corr\,($\uparrow$)\\ \midrule
\multirow{5}{*}{\makecell{Similarity-\\\!\!\!\!\!\!\!\!\!\!\!\!based}} & ROUGE-1 & Not Available & 0.324 (0.00) & 0.332 (0.00) & 0.883 (0.00)\\
 & ROUGE-2 & Not Available & 0.384 (0.00) & 0.315 (0.00) & 0.947 (0.00)\\
 & ROUGE-L & Not Available & 0.175 (0.00) & 0.180 (0.00) & 0.667 (0.05)\\
 & BERTScore & Not Available & 0.008 (0.69) & 0.000 (0.97)  & -0.133 (0.73) \\
 & BARTScore & Not Available & 0.717 (0.00) & 0.736 (0.00)  & 0.937 (0.00) \\ \midrule
NLI-based & SummaC-Conv & Not Available & 0.828 (0.00) & 0.814 (0.00) & 0.883 (0.00) \\ \midrule
\multirow{2}{*}{QA-based} & UniEval & Not Available & 0.743 (0.00) & 0.772 (0.00)  &  0.983 (0.00) \\
 & QAFactEval & Not Available & 0.841 (0.00) & 0.813 (0.00)  & 0.933 (0.00) \\ \midrule

\multirow{2}{*}{LLM-based} & G-Eval\,(GPT-4) & Not Available & \textbf{0.841} (0.00) & 0.834 (0.00)  & \textbf{0.950} (0.00) \\
 & \textbf{\algname{}\,(GPT-4) } & \textbf{86.4\%} & 0.833 (0.00) & \textbf{0.839} (0.00) & \textbf{0.950} (0.00) \\ \bottomrule
\end{tabular}
\end{center}
\vspace*{-0.4cm}
\caption{Performance of \textbf{faithfulness evaluation on FRANK} using ten automated metrics at the sentence-, summary- and system-level. The values in parenthesis represent p-values. The best results are marked in bold.} 
\label{table:table_faithfulness}
\vspace*{-0.4cm}
\end{table*}

\subsection{Keyfact Extraction}

The list of keyfacts is essential for evaluating the completeness and conciseness using \algname{}. Humans are best suited to generate these keyfacts as they understand the priorities in different domains, such as medicine or sales. However, in some cases, obtaining human keyfacts can be challenging.
\algname{} works with human keyfacts by default, but for cases where no human keyfacts are provided, it can employ the LLM to extract keyfacts automatically. This process is entirely automated, utilizing prompts tailored for keyfact extraction (see Figure \ref{fig:keyfact_extract}). For further details, refer to Appendix \ref{sec:deatail_keyfact_extraction}. The impact of employing automatic keyfact extraction on keyfact alignment is discussed in Section \ref{sec:eval_others}.

\section{Evaluation}\label{sec:exp}
\vspace*{-0.05cm}

\paragraph{Datasets} 

To evaluate the automated evaluator's performance, we need datasets with human annotations for sentence-level faithfulness and keyfact-level alignment. Since \emph{no} single dataset includes both types of annotations, we opt for two separate datasets. FRANK\,\cite{pagnoni2021understanding} is a benchmark dataset of $2,246$ summaries for factuality evaluation metrics. It encompasses summaries of nine summarization systems on CNNDM\,\cite{hermann2015teaching} and XSUM\,\cite{Narayan2018DontGM}, providing fine-grained annotations of sentence-level factuality error types. On the other hand, REALSumm\,\cite{bhandari2020re} is another dataset of $2,500$ summaries from 25 summarization systems for automated metrics based on CNNDM. It includes a list of human keyfacts, along with corresponding annotations indicating their presence in the summary. FRANK and REALSumm obtain the inter-annotator agreement\,(IAA) scores of 0.58 (cohen's kappa) and 0.66 (Krippendorff's alpha) for three annotators, respectively.

\paragraph{LLMs as Evaluators} 
We use the GPT-4-turbo (gpt-4-1106-preview) \cite{achiam2023gpt} by default in main evaluation, but test with various open-source and proprietary LLMs, including Mixtral-8x7B\,\cite{jiang2024mixtral}, Phi-2, Llama-2/-3\,\cite{touvron2023llama}, GPT-3.5-turbo, and GPT-4-omni (gpt-4o-2024-05-13), in Section \ref{sec:exp_many_llms}. We set the temperature to 0 and clear the history for every evaluation instance, following the literature\,\cite{shen2023large}. We use HuggingFace models for open-source LLMs and paid APIs for proprietary LLMs.

\paragraph{Baselines} 
We compare \algname{} with five similarity-based methods, ROUGE-1/-2/-L \cite{lin2004rouge}, BERTScore \cite{zhang2019bertscore}, and BARTScore \cite{yuan2021bartscore}; a NLI-based method, SummaC-Conv \cite{laban2022summac}; two QA-based methods, UniEval \cite{zhong2022towards} and QAFactEval \cite{fabbri2022qafacteval}; and the latest LLM-based method, G-Eval \cite{liu2023gpteval}. Note that QAFactEval and SummaC-Conv are only compared for faithfulness evaluation, as they are limited to factuality. We obtain all the results by executing each metric in our experimental setup.

\paragraph{Performance} Each automated evaluator's performance is assessed by comparing estimated scores with ground-truth human judgments using \emph{sentence}, \emph{summary}, and \emph{system}-level measurements. 
This multi-level analysis is crucial, as we seek to understand the agreement of the evaluator on each sentence, each summary, and the average performance of each summarization system.

\emph{Balanced accuracy (bAcc)} assesses faithfulness in classifying each summary sentence for the presence or absence of factual errors at the sentence-level. This is the average of true positive and true negative rates widely used when the two classes are imbalanced\,\cite{brodersen2010balanced}. 
\emph{Pearson} and \emph{Spearman correlations} assess all three dimensions at the summary-level by comparing percentage scores in Eqs.\,\eqref{eq:faithfulness_score}-\eqref{eq:other_scores} derived from predicted and human evaluation results.
Lastly, \emph{rank correlation} is a system-level measure assessing the alignment of performance rankings across summarization systems (models) calculated by both our evaluator and humans.
The details of the measurements are provided in Appendix \ref{sec:measurement}.

\begin{table*}[t]
\begin{center}
\footnotesize
\begin{tabular}{|L{3.3cm} |L{1.1cm} X{1.1cm} X{1.1cm} X{1.1cm} X{1.1cm} X{1.1cm} X{1.1cm} |X{1.1cm}|}\toprule
Error Category & OutE & EntE & PredE & CirE & GramE & LinkE & CorefE  & Mean  \\ \midrule
Random Guessing & 14.3\% & 14.3\% & 14.3\% & 14.3\% & 14.3\% & 14.3\% & 14.3\% & 14.3\% \\
Bart-Large (Fine-tuned) & \textbf{56.7}\% & 36.9\% & 14.8\% & 34.0\% & 21.4\% & 0.0\% & \textbf{40.0}\% & 29.1\% \\\midrule
\textbf{\algname{}\,(GPT-4)} & 50.2\% & \textbf{63.7}\% & \textbf{41.9}\% & \textbf{38.1}\% & \textbf{44.6}\% & \textbf{19.4}\% & {37.8\%} & \textbf{42.2}\% \\ \bottomrule
\end{tabular}
\end{center}
\vspace*{-0.4cm}
\caption{Accuracy analysis of \textbf{factuality error localization} in assessing faithfulness, with error categories including OutE\,(out-of-context), EntE\,(entity error), PredE\,(predicate error), CirE\,(circumstance error), GramE (grammatical error), LinkE\,(discourse link error), and CorefE\,(coreference error). "Random Guessing" is the performance of randomly selecting from the seven categories, i.e., 1/7=14.3\%, while "Bart-Large" is a stronger baseline model fine-tuned on FRANK for error localization. } 
\label{table:main_error_type}
\vspace*{-0.1cm}
\end{table*}


\begin{table*}[t]
\begin{center}
\footnotesize
\begin{tabular}{|L{1.5cm} | L{2.5cm} |X{1.4cm} X{1.4cm} X{1.5cm} |X{1.4cm} X{1.4cm} X{1.5cm}|}\toprule
\multicolumn{2}{|c|}{Dimension}  & \multicolumn{3}{c|}{(a) Completeness} & \multicolumn{3}{c|}{(b) Conciseness}  \\\midrule
Direction & Method &  \multicolumn{2}{c}{Summary-level} & \!\!\!System-level\!\!\! & \multicolumn{2}{c}{Summary-level} & \!\!\!System-level\!\!\!  \\ 
 &  & \!\!\!Pearson\,($\uparrow$)\!\!\! & \!\!\!Spearman\,($\uparrow$)\!\!\! & Rank\,($\uparrow$) & \!\!\!Pearson\,($\uparrow$)\!\!\! & \!\!\!Spearman\,($\uparrow$)\!\!\! & Rank\,($\uparrow$) \\ \midrule
\multirow{5}{*}{\makecell{Similarity-\\\!\!\!\!\!\!\!\!\!\!\!\!based}} & ROUGE-1 & \!\!\!0.484 (0.00)\!\!\! & \!\!\!0.461 (0.00)\!\!\! & \!\!\!0.516 (0.01)\!\!\! & \!\!\!0.387 (0.00)\!\!\! & \!\!\!0.371 (0.00)\!\!\! & \!\!\!0.332 (0.10)\!\!\! \\
& ROUGE-2 & \!\!\!0.456 (0.00)\!\!\! & \!\!\!0.461 (0.00)\!\!\! & \!\!\!0.463 (0.02)\!\!\! & \!\!\!0.328 (0.00)\!\!\! & \!\!\!0.337 (0.00)\!\!\! & \!\!\!0.290 (0.16)\!\!\! \\
& ROUGE-L & \!\!\!0.425 (0.00)\!\!\! & \!\!\!0.428 (0.00)\!\!\! & \!\!\!0.238 (0.25)\!\!\! & \!\!\!0.310 (0.00)\!\!\! & \!\!\!0.321 (0.00)\!\!\! & \!\!\!0.083 (0.69)\!\!\! \\
& BERTScore & \!\!\!0.455 (0.00)\!\!\! & \!\!\!0.443 (0.00)\!\!\! & \!\!\!0.619 (0.00)\!\!\! & \!\!\!0.416 (0.00)\!\!\! & \!\!\!0.405 (0.00)\!\!\! & \!\!\!0.783 (0.00)\!\!\! \\
& BARTScore & \!\!\!0.216 (0.00)\!\!\! & \!\!\!0.199 (0.00)\!\!\! & \!\!\!0.653 (0.00)\!\!\! & \!\!\!0.241 (0.00)\!\!\! & \!\!\!0.210 (0.00)\!\!\! & \!\!\!0.824 (0.00)\!\!\! \\\midrule
QA-based & UniEval & \!\!\!0.134 (0.00)\!\!\! & \!\!\!0.180 (0.00)\!\!\! & \!\!\!0.346 (0.09)\!\!\! & \!\!\!0.086 (0.00)\!\!\! & \!\!\!0.128 (0.00)\!\!\! & \!\!\!-0.176 (0.39)\!\!\! \\ \midrule
\multirow{3}{*}{LLM-based}& G-Eval\,(GPT4) & \!\!\!0.314 (0.00)\!\!\! & \!\!\!0.295 (0.00)\!\!\! & \!\!\!0.908 (0.00)\!\!\! & \!\!\!0.314 (0.00)\!\!\! & \!\!\!0.277 (0.00)\!\!\! & \!\!\!0.582 (0.00)\!\!\! \\
& \textbf{\algname{}\,(GPT-4)}\!\! & \!\!\!\textbf{0.688} (0.00)\!\!\! & \!\!\!\textbf{0.677} (0.00)\!\!\! & \!\!\!\textbf{0.949} (0.00)\!\!\! & \!\!\!\textbf{0.505} (0.00)\!\!\! & \!\!\!\textbf{0.451} (0.00)\!\!\! & \!\!\!{0.880} (0.00)\!\!\! \\ 
& {$\text{\algname{}}^{\dagger}$(GPT-4)}\!\! & \!\!\!{0.571} (0.00)\!\!\! & \!\!\!{0.546} (0.00)\!\!\! & \!\!\!{0.905} (0.00)\!\!\! & \!\!\!{0.438} (0.00)\!\!\! & \!\!\!{0.399} (0.00)\!\!\! & \!\!\!\textbf{0.911} (0.00)\!\!\! \\ \bottomrule
\end{tabular}
\end{center}
\vspace*{-0.45cm}
\caption{Performance of \textbf{completeness and conciseness evaluation on REALSumm} using ten automated evaluation metrics at the summary- and system-level. The values in parenthesis represent p-values. $\text{Fine-Eval}^{\dagger}$ utilizes the list of keyfacts automatically derived through LLMs, in contrast to relying on human keyfacts.} 
\label{table:table_keyfact}
\vspace*{-0.35cm}
\end{table*}

\subsection{Main Results: Evaluators Comparison}


\subsubsection{Faithfulness}
\label{sec:eval_faithfulness}


\begin{table}[t]
\begin{center}
\footnotesize
\begin{tabular}{|L{1.2cm} |X{1.57cm} X{1.57cm} X{1.57cm}|}\toprule
Method & Faithfulness & \!\!\!\!Completeness\!\!\!\! & Conciseness \\ \midrule
\!G-Eval\!\! & 0.906 & 0.799 & 0.759 \\ 
\!\textbf{\algname{}}\!\! & \textbf{0.921} & \textbf{0.853} & \textbf{0.908} \\ \bottomrule
\end{tabular}
\end{center}
\vspace*{-0.45cm}
\caption{\textbf{Inter-annotator agreement score\,(IAA)} of GEval and \algname{} across three distinct evaluations.} 
\label{table:stability}
\vspace*{-0.5cm}
\end{table}

Table \ref{table:table_faithfulness} summarizes the agreement between automated evaluators and human scores in faithfulness evaluation at three different granularities.
\algname{} significantly outperforms similarity-, NLI-, and QA-based evaluators at all levels of evaluation. 

It is important to note that none of the existing methods provide sentence-level evaluation results, relying instead on summary-level scoring, such as Likert-scale scores. It is noteworthy that \algname{} has the capability to assess whether each sentence contains a factual error or not, demonstrating remarkable alignment with human sentence-level judgments, with a balanced accuracy of 86.4\%. 

\footnotetext[3]{The pre-trained Bart-Large\,\cite{lewis2020bart} is fine-tuned on error localization data constructed using FRANK, comprising 3,885 training and 1,007 testing sentences, each paired with their corresponding human error categories.}

Given the strong alignment with human judgment, using LLMs as an evaluator holds great promise for enhancing the reliability of evaluation processes for text summarization. 
However, one open question remains: \emph{Can LLMs identify the type of factuality error?}

Table \ref{table:main_error_type} unveils the capability of LLMs for factuality error localization, demonstrating accuracy as the probability that the predicted error category matches the correct answer given by humans.
\algname{} outperforms the strong baseline, Bart-Large\footnotemark[3] fine-tuned on FRANK for error localization, despite not being trained on any error localization data, i.e., zero-shot prediction. Its superiority is primarily stemming from error categories that are uncommon in the training set for Bart-Large, such as PredE (141 cases), CirE (142 cases), and LinkE (41 cases).
Nevertheless, LLMs still make numerous mistakes in accurately identifying the exact error type, despite their excellent performance in the binary decision of hallucination.

Therefore, achieving a level of evaluation comparable to human performance in more intricate assessment tasks remains a challenging objective.

\begin{table*}[t]
\begin{center}
\footnotesize
\begin{tabular}{|L{0.7cm} |L{2.4cm} |X{2.1cm} |X{2.15cm} X{2.15cm} X{2.15cm}| X{1.35cm}| }\toprule
Type & LLM & Sentence-level & \multicolumn{2}{c}{Summary-level} & System-level & \!\!\!Success\!\!\! \\
 &  & bAcc\,($\uparrow$) & \!\!Pearson Corr\,($\uparrow$)\!\! & \!\!Spearman Corr\,($\uparrow$)\!\! & Rank Corr\,($\uparrow$) & Ratio\\ \midrule
\multirow{6}{*}{\hspace*{0.15cm}\rotatebox[origin=c]{90}{\makecell{\footnotesize ~~Open-source}}}& \!Phi-2 (2.7B) & 48.1\% & \!\!-0.108 (0.00) & \!\!-0.010 (0.73) & \!\!-0.700 (0.04) & 50.4\% \\
&\!Llama2-70B & 56.5\% & 0.133 (0.00) & 0.147 (0.00) & 0.833 (0.01) & 86.2\% \\
&\!Mixtral-8x7B & 50.7\% & -0.023 (0.38) & 0.036 (0.18) & \!\!-0.450 (0.22) & 63.1\% \\ 
&\!Mixtral-8x7B-Inst.\!\! & 78.7\% & 0.708 (0.00) & 0.716 (0.00) & 0.883 (0.00) & 88.9\% \\ 
&\!Llama3-70B-Inst. & \textbf{92.0}\% & {0.844} (0.00) & {0.841} (0.00) & 0.933 (0.00) & \textbf{98.3}\% \\ 
&\!Gemma2-27B-Inst. & {90.8}\% & {0.838} (0.00) & {0.833} (0.00) & \textbf{0.950} (0.00) & {97.8}\% \\ \midrule
\multirow{4}{*}{\hspace*{0.15cm}\rotatebox[origin=c]{90}{\makecell{\footnotesize Proprietary}}} 
&\!Gemini-1-pro & 87.7\% & 0.733 (0.00) & 0.736 (0.00) &  0.916 (0.00) & 98.0\% \\ 
&\!GPT-3.5-turbo& 78.8\% & 0.709 (0.00) & 0.709 (0.00) & 0.933 (0.00) & 93.1\% \\
&\!GPT-4-turbo& {86.4\%} & {0.833} (0.00) & {0.839} (0.00) & \textbf{0.950} (0.00) & 98.1\% \\
&\!GPT-4-omni& {91.8\%} & \textbf{0.855} (0.00) & \textbf{0.852} (0.00) & {0.883} (0.00) & 98.1\% \\ \bottomrule
\end{tabular}
\end{center}
\vspace*{-0.45cm}
\caption{Performance of \textbf{faithfulness evaluation} using six open-source and four proprietary LLMs. The rightmost column is the success ratio of accurately following the prompt.} 
\label{table:llm_faithfulness}
\vspace*{-0.4cm}
\end{table*}

\subsubsection{Completeness and Conciseness}
\label{sec:eval_others}

The agreement between automated evaluators and human scores on completeness and conciseness is summarized in Table \ref{table:table_keyfact}. In contrast to similarity-based evaluators, which provide a single composite score, UniEval and G-Eval yield four distinct scores, evaluating faithfulness, coherence, relevance, and fluency. We use their coherence and relevance scores to calculate the correlation with human scores for completeness and conciseness, as they indicate the inclusion and density of key information, respectively. 

Overall, \algname{} using human keyfacts demonstrates a very high agreement with human evaluations for completeness and conciseness, surpassing other evaluators significantly.
This is because keyfact alignment is essential to verify the coverage of crucial information in the summary, a task that cannot be accomplished with existing LLM-based method like G-Eval. See the qualitative example in Appendix \ref{sec:qualitatitve_completeness}.
We also assess the performance of \algname{} without employing human keyfacts and, instead, utilizing machine-generated keyfacts, as outlined in Appendix \ref{sec:deatail_keyfact_extraction}. The keyfacts are extracted using GPT-4 with a specific prompt. It is noteworthy that, even with machine-generated key facts, \algname{} maintains a higher level of agreement over other automated evaluators.

With an advantage as a fine-grained evaluator, \algname{} also provides evaluation results at the keyfact-level, revealing which keyfacts are omitted in the summary, i.e., keyfact matching. Given a list of keyfacts, it includes binary labels ("Yes" or "No") in the JSON output, as illustrated in Figure \ref{fig:best_prompt_keyfact}. 
Therefore, we assess the agreement for the keyfact matching task by calculating the IAA score between machine and human labels. \algname{} demonstrates a Krippendorff's alpha of 0.65 for keyfact matching. 
This robust agreement at various levels corroborates that \algname{} has a potential to be an effective fine-grained automatic evaluator.

Furthermore, in Appendix \ref{sec:fair_g_eval}, we compare \algname{} with two variants of G-Eval, which are tailored for completeness and conciseness evaluation by modifying its prompts to be more suitable for such assessment and integrating them for use with keyfacts. \algname{} maintains its significant dominance even with additional tuning on G-Eval.

\begin{table*}[t]
\begin{center}
\footnotesize
\begin{tabular}{|L{0.7cm} | L{2.3cm} |X{1.4cm} X{1.4cm} X{1.5cm} |X{1.4cm} X{1.4cm} X{1.5cm}| X{0.6cm}| }\toprule
\multicolumn{2}{|c|}{Dimension}  & \multicolumn{3}{c|}{(a) Completeness} & \multicolumn{3}{c|}{(b) Conciseness} &   \\\midrule
Type& Method &  \multicolumn{2}{c}{Summary-level} & \!\!\!System-level\!\!\! & \multicolumn{2}{c}{Summary-level} & \!\!\!System-level\!\!\! & \!\! Succ.\!\!  \\ 
&  & \!\!\!Pearson\,($\uparrow$)\!\!\! & \!\!\!Spearman\,($\uparrow$)\!\!\! & Rank\,($\uparrow$) & \!\!\!Pearson\,($\uparrow$)\!\!\! & \!\!\!Spearman\,($\uparrow$)\!\!\! & Rank\,($\uparrow$) & \!\!Ratio\!\! \\ \midrule
\multirow{6}{*}{\hspace*{0.15cm}\rotatebox[origin=c]{90}{\makecell{\footnotesize ~~Open-source}}}& \!Phi-2 (2.7B) & \!\!\!0.093 (0.00)\!\!\! & \!\!\!0.104 (0.00)\!\!\! & \!\!\!\!\!0.338 (0.10)\!\!\! & \!\!\!0.058 (0.04)\!\!\! & \!\!\!0.069 (0.01)\!\!\! & \!\!\!\!\!-0.039 (0.85)\!\!\! & \!\!\!52.1\%\!\!\! \\
&\!Llama2-70B & \!\!\!0.421 (0.00)\!\!\! & \!\!\!0.401 (0.00)\!\!\! & \!\!\!\!\!0.824 (0.00)\!\!\! & \!\!\!0.387 (0.00)\!\!\! & \!\!\!0.371 (0.00)\!\!\! & \!\!\!\!0.612 (0.00)\!\!\! & \!\!\!53.7\%\!\!\! \\
&\!Mixtral-8x7B & \!\!\!0.166 (0.00)\!\!\! & \!\!\!0.152 (0.00)\!\!\! & \!\!\!0.431 (0.03)\!\!\! & \!\!\!0.087 (0.00)\!\!\! & \!\!\!0.108 (0.00)\!\!\! & \!\!\!0.264 (0.20)\!\!\! &  \!\!\!53.8\%\!\!\! \\
&\!Mixtral-8x7B-Inst.\!\! & \!\!\!0.439 (0.00)\!\!\! & \!\!\!0.437 (0.00)\!\!\! & \!\!\!0.678 (0.00)\!\!\! & \!\!\!0.367 (0.00)\!\!\! & \!\!\!0.361 (0.00)\!\!\! & \!\!\!0.798 (0.00)\!\!\! & \!\!\!87.5\%\!\!\! \\ 
&\!Llama3-70B-Inst.\!\! & \!\!\!\textbf{0.755} (0.00)\!\!\! & \!\!\!\textbf{0.747} (0.00)\!\!\! & \!\!\!\!\!0.881 (0.00)\!\!\! & \!\!\!0.445 (0.00)\!\!\! & \!\!\!0.444 (0.00)\!\!\! & \!\!\!\!0.786 (0.00)\!\!\! & \!\!\!92.0\%\!\!\! \\
&\!Gemma2-27B-Inst.\!\!\!\! & \!\!\!{0.585} (0.00)\!\!\! & \!\!\!{0.568} (0.00)\!\!\! & \!\!\!\!\!0.869 (0.00)\!\!\! & \!\!\!0.434 (0.00)\!\!\! & \!\!\!0.397 (0.00)\!\!\! & \!\!\!\!0.850(0.00)\!\!\! & \!\!\!\textbf{99.8}\%\!\!\! \\\midrule
\multirow{4}{*}{\hspace*{0.15cm}\rotatebox[origin=c]{90}{\makecell{\footnotesize Proprietary}}} 
&\!Gemini-1-pro & \!\!\!0.583 (0.00)\!\!\! & \!\!\!0.567 (0.00)\!\!\! & \!\!\!0.820 (0.00)\!\!\! & \!\!\!0.435 (0.00)\!\!\! & \!\!\!0.402 (0.00)\!\!\! & \!\!\!0.745 (0.00)\!\!\! & \!\!\!99.7\%\!\!\! \\
&\!GPT-3.5-turbo & \!\!\!0.509 (0.00)\!\!\! & \!\!\!0.493 (0.00)\!\!\! & \!\!\!0.848 (0.00)\!\!\! & \!\!\!0.381 (0.00)\!\!\! & \!\!\!0.372 (0.00)\!\!\! & \!\!\!0.706 (0.00)\!\!\! & \!\!\!74.5\%\!\!\! \\
&\!GPT-4-turbo& \!\!\!{0.688} (0.00)\!\!\! & \!\!\!{0.677} (0.00)\!\!\! & \!\!\!{0.949} (0.00)\!\!\! & \!\!\!{0.505} (0.00)\!\!\! & \!\!\!{0.451} (0.00)\!\!\! & \!\!\!{0.880} (0.00)\!\!\! & \!\!\!\textbf{99.8}\%\!\!\!  \\ 
&\!GPT-4-omni& \!\!\!{0.691} (0.00)\!\!\! & \!\!\!{0.686} (0.00)\!\!\! & \!\!\!\textbf{0.943} (0.00)\!\!\! & \!\!\!\textbf{0.522} (0.00)\!\!\! & \!\!\!\textbf{0.467} (0.00)\!\!\! & \!\!\!\textbf{0.932} (0.00)\!\!\! & \!\!\!{99.6}\%\!\!\!  \\ \bottomrule
\end{tabular}
\end{center}
\vspace*{-0.45cm}
\caption{Performance of \textbf{completeness and conciseness} evaluation using six open-source and four proprietary LLMs. The rightmost column is the success ratio of accurately following the prompt.} 
\label{table:llm_keyfact}
\vspace*{-0.10cm}
\end{table*}

\begin{table*}[t]
\begin{center}
\footnotesize
\begin{tabular}{|L{0.7cm}  |L{2.5cm} |L{1.06cm} X{1.06cm} X{1.06cm} X{1.06cm} X{1.06cm} X{1.06cm} X{1.06cm} |X{1.06cm}|}\toprule
Type & LLM  & \multicolumn{7}{c|}{Factuality Error Type} &  \\
& Models & OutE & EntE & PredE & CirE & GramE & LinkE & CorefE  & Mean  \\ \midrule
\multicolumn{2}{|c|}{Baseline (Random Guessing)} & 14.3\% & 14.3\% & 14.3\% & 14.3\% & 14.3\% & 14.3\% & 14.3\% & 14.3\% \\\midrule
\multirow{6}{*}{\hspace*{0.15cm}\rotatebox[origin=c]{90}{\makecell{\footnotesize ~~Open-source}}} & Phi-2 & ~~8.1\% & 12.6\% & ~~4.5\% & ~~5.8\% & ~~6.9\% & ~~8.3\% & ~~6.8\% & ~~7.6\% \\
& Llama2-70B & 21.6\% & 31.4\% & 13.4\% & 14.3\% & 19.0\% & 11.1\% & 25.0\% & 19.4\% \\
& Mixtral-8x7b & 21.8\% & 24.1\% & 16.0\% & ~~6.7\% & 15.6\% & ~~7.1\% & 5.6\% & 13.8\% \\
& Mixtral-8x7b-Inst. & 37.8\% & 45.4\% & 26.1\% & 12.5\% & 22.2\% & {25.0}\% & 22.7\% & 27.4\% \\
& Llama3-70B-Inst. & 66.1\% & 64.8\% & 41.1\% & 38.1\% & \textbf{54.5}\% & \textbf{43.8}\% & 37.5\% & 49.4\% \\ 
& Gemma2-27B-Inst. & 63.1\% & 64.7\% & 44.5\% & \textbf{47.8}\% & {43.0}\% & {19.1}\% & 30.0\% & 44.6\% \\ \midrule
\multirow{4}{*}{\hspace*{0.15cm}\rotatebox[origin=c]{90}{\makecell{\footnotesize Proprietary}}} 
& Gemini-1-pro & 51.9\% & 36.0\% & {23.2\%} & 18.2\% & ~~3.8\% & {0.0}\% & 25.0\% & 22.6\% \\
& GPT-3.5-turbo & 52.4\% & 42.4\% & 26.4\% & 25.9\% & 52.4\% & 0.0\% & 12.9\% & 30.3\% \\
& GPT-4-turbo & 50.2\% & {63.7}\% & 41.9\% & 38.1\% & 44.6\% & 19.4\% & {37.8\%} & 42.2\% \\ 
& GPT-4-omni & \textbf{70.6}\% & \textbf{69.2}\% & \textbf{45.7\%} & {46.6}\% & 50.0\% & {42.3}\% & \textbf{44.0}\% & \textbf{52.6}\% \\\bottomrule
\end{tabular}
\end{center}
\vspace*{-0.4cm}
\caption{Accuracy analysis of \textbf{factuality error localization} in assessing faithfulness using six open-source and four proprietary LLMs, where "Baseline" is the performance of random guessing. Top-1 values are marked in bold.} 
\label{table:llm_fact_error}
\vspace*{-0.4cm}
\end{table*}

\subsubsection{Stability in Evaluation Results}
\label{sec:stability}

Concerns arise about evaluation result stability with LLMs due to their inherent text generation randomness, even at temperature 0.
Despite LLM-based methods relying on Likert-scale evaluation, such as G-Eval, showing significant fluctuations in judgment alignment\,\cite{shen2023large, liu2023gpteval}, Table \ref{table:stability} demonstrates that \algname{} (GPT-4) maintains much higher agreement in summary-level evaluation scores across three distinct runs.
This underscores the benefit of employing fine-grained percentage scores derived from sentence- and keyfact-level assessments.

\subsection{LLMs as Evaluators Comparison}
\label{sec:exp_many_llms}

It is interesting to observe how the evaluation agreement varies based on the choice of LLMs, given the abundance of open-source and proprietary LLMs. 

\vspace*{-0.1cm}
\paragraph{Success Ratio.}
\label{sec:success_ratio}
The primary limitation of open-source LLMs is their comparatively lower success ratio in following prompts, compared to proprietary LLMs; only Llama3-70B-Inst exhibits a high success ratio comparable to proprietary LLMs. Upon analyzing failure cases, the top three reasons are: (1) the output is either not in JSON format or an incorrect JSON format, (2) the output consists of meaningless text, e.g., python codes or no output at all, and (3) the JSON output includes only a few lines of sentences or keyfacts. 

Furthermore, the maximum token length in context for open-source LLMs is notably shorter compared to proprietary LLMs. GPT-4 series can process up to 128K  tokens, whereas open-source LLMs generally handle up to 8K input tokens. This results in prompt truncation when handling lengthy input texts, potentially leading to failures in generating accurate outputs in text summarization.

\paragraph{Agreement with Human Score.}

Tables \ref{table:llm_faithfulness}-\ref{table:llm_keyfact} summarize the correlation of nine different LLMs with human judgment, computed only for the successful cases of adhering to the prompt.
Although the recent Llama3-70B-Inst shows strong agreement with humans, in general, there is a noticeable gap  between open-source and proprietary LLMs. 
Regarding open-source LLMs, the agreement with human scores increases with the model size; for example, Llama2-70B exhibits a higher correlation coefficient than Phi-2.
Additionally, instruction tuning also plays a role, as observed in Mixtral-8x7b's performance, which improved significantly after instruction tuning. In contrast, all the proprietary LLMs exhibit high correlation coefficient. Particularly, more recent and powerful LLMs exhibit better performance, i.e., GPT-4-turbo > GPT-3.5-turbo, GPT-4-omni > GPT-4-turbo.

It's notable that LLMs with a high success ratio exhibit a strong correlation, suggesting they are not penalized by their high success ratios. Therefore, a more advanced LLM simultaneously achieves higher agreement and success ratios.

\begin{figure*}[t!]
\begin{center}
\includegraphics[width=16.1cm]{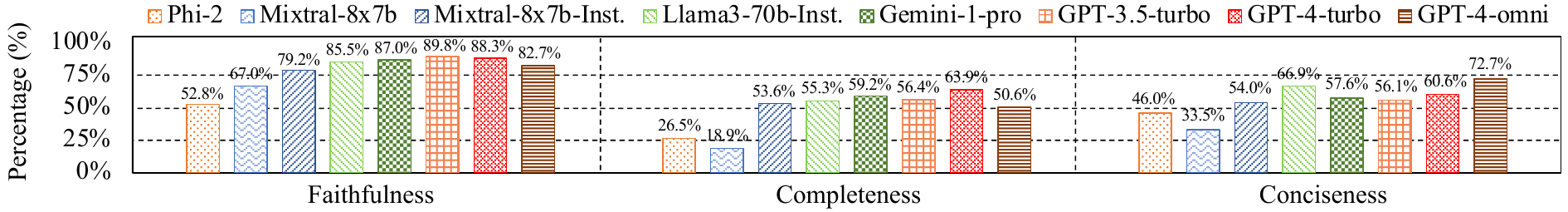}
\end{center}
\vspace*{-0.35cm}
\caption{\textbf{Evaluation using \algname{}} for eight LLMs in text summarization on CNNDM. }
\vspace*{-0.4cm}
\label{fig:soto_llm_eval}
\end{figure*}

\paragraph{Error Localization}
We provide a detailed factuality error localization analysis using different LLMs in Table \ref{table:llm_fact_error}.
GPT-4-omni improves the mean accuracy in error localization by 10\% over GPT-4-turbo. 
The categorization accuracies of open-source LLMs are considerably lower than those of proprietary LLMs in general. However, the latest open-source LLM, Llama3-70B-Inst, outperforms GPT-4-turbo in error localization, achieving an average prediction accuracy of $49.4\%$, which is $7.2\%$ higher than that of GPT-4-turbo.
Additionally, instruction tuning demonstrates a significant accuracy boost in this task, as evidenced by the improvement from Mixtral-8x7b to Mixtral-8x7b-Inst. 

\vspace*{-0.1cm}
\subsection{Evaluation using \algname{}}
\vspace*{-0.1cm}

As an actual application of an automated evaluator, we gather summaries generated by four open-source and four proprietary LLMs, and subsequently assess their summarization quality using the \algname{} algorithm (see the prompt for summarization we used in Appendix \ref{sec:llm_summary}). Figure \ref{fig:soto_llm_eval} shows the percentage scores of the eight LLMs for faithfulness, completeness, and conciseness. The summaries are generated for 100 examples sourced from CNNDM, all of which are also included in REALSumm, thereby possessing the list of keyfacts extracted by human annotators.

In general, proprietary LLMs, including different versions of GPT, generate high-quality summaries in comparison to open-source counterparts. Interestingly, GPT-4-omni exhibits the highest agreement with humans as an automated evaluator in Tables \ref{table:llm_faithfulness}-\ref{table:llm_keyfact}, but its faithfulness and completeness scores are significantly worse even than GPT-3.5-turbo. Consequently, GPT-4-omni is likely to include more hallucinations and miss many important keyfacts in summary generation.

The performance ranking of each model changes significantly for each evaluation dimension. GPT-3.5-turbo, GPT-4-turbo, and GPT-4-omni are the best for faithfulness, completeness, and conciseness, respectively.
Nevertheless, it is noteworthy that Llama3-70B-Inst, an open-source LLM, exhibits comparable performance to the state-of-the-art proprietary LLMs. 
In open-source LLMs, instruction tuning significantly enhances summarization quality, as evidenced by the performance increase of Mixtral-8x7b-Inst over Mixtral-8x7b. 
These findings align with prior observations reported in recent studies on faithfulness \cite{laban2023summedits} and instruction tuning \cite{zhang2023instruction}. 

Lastly, while there is no doubt that faithfulness is crucial, achieving both completeness and conciseness simultaneously turns out to be very important and challenging in text summarization, as evident from the low percentage scores even with GPT-4 series. Therefore, it emphasizes the need to put more effort into these aspects for a good summary.

\section{Conclusion}\label{sec:con}

%
%

We introduce \algname{}, a novel automated evaluator designed for fine-grained and multi-dimensional text summarization evaluation. The evaluation process is broken down into fact checking and keyfact alignment, providing detailed insights, where keyfacts can be either provided by humans or generated by LLMs. Our experiments include a thorough comparison with existing evaluators, exploration of performance using eight opensource or proprietary LLMs, and real quality assessment of recent LLM-generated summaries. The results indicate the potential effectiveness of \algname{} as a text summarization evaluator, showcasing its promising capabilities in advancing automated evaluation for text summarization.

\section*{Limitations} 
Our automated evaluator is primarily tested on the news domain due to the limited availability of benchmark datasets with fine-grained human annotations. We emphasize the critical importance of constructing a high-quality benchmark dataset with high diversity in input domains, length, and types. Also, the prompts for evaluation may need to be tuned if a different summary is expected like the summary from the medical domain. Lastly, other aspects can be considered for text summarization, such as toxicity and social bias. We leave these challenges as future work.

\section*{Ethics Statement}
This paper focuses on designing an automatic evaluator using LLMs for text summarization. Therefore, we do not anticipate any negative ethical and social impact. 

\section*{Acknowledgements}
The first author, Hwanjun Song, was supported by the National Research Foundation of Korea (NRF) grant funded by the Korea government (MSIT) (No. RS-2024-00334343) and Artificial Intelligence industrial convergence cluster development project funded by the Ministry of Science and ICT (MSIT, Korea) \& Gwangju Metropolitan City (No. BA00000772).



\begin{thebibliography}{49}
\expandafter\ifx\csname natexlab\endcsname\relax\def\natexlab#1{#1}\fi

\bibitem[{Achiam et~al.(2023)Achiam, Adler, Agarwal, Ahmad, Akkaya, Aleman, Almeida, Altenschmidt, Altman, Anadkat et~al.}]{achiam2023gpt}
Josh Achiam, Steven Adler, Sandhini Agarwal, Lama Ahmad, Ilge Akkaya, Florencia~Leoni Aleman, Diogo Almeida, Janko Altenschmidt, Sam Altman, Shyamal Anadkat, et~al. 2023.
\newblock G{PT}-4 technical report.
\newblock \emph{arXiv preprint arXiv:2303.08774}.

\bibitem[{Banerjee and Lavie(2005)}]{banerjee2005meteor}
Satanjeev Banerjee and Alon Lavie. 2005.
\newblock {METEOR}: An automatic metric for mt evaluation with improved correlation with human judgments.
\newblock In \emph{ACLW}.

\bibitem[{Bhandari et~al.(2020)Bhandari, Gour, Ashfaq, Liu, and Neubig}]{bhandari2020re}
Manik Bhandari, Pranav~Narayan Gour, Atabak Ashfaq, Pengfei Liu, and Graham Neubig. 2020.
\newblock Re-evaluating evaluation in text summarization.
\newblock In \emph{EMNLP}.

\bibitem[{Brodersen et~al.(2010)Brodersen, Ong, Stephan, and Buhmann}]{brodersen2010balanced}
Kay~Henning Brodersen, Cheng~Soon Ong, Klaas~Enno Stephan, and Joachim~M Buhmann. 2010.
\newblock The balanced accuracy and its posterior distribution.
\newblock In \emph{ICPR}.

\bibitem[{Chen et~al.(2023)Chen, Wang, Jiang, Shi, and Xu}]{chen2023exploring}
Yi~Chen, Rui Wang, Haiyun Jiang, Shuming Shi, and Ruifeng Xu. 2023.
\newblock Exploring the use of large language models for reference-free text quality evaluation: A preliminary empirical study.
\newblock \emph{arXiv preprint arXiv:2304.00723}.

\bibitem[{Chen et~al.(2021)Chen, Liu, and Qiu}]{chen2021factuality}
Yiran Chen, Pengfei Liu, and Xipeng Qiu. 2021.
\newblock Are factuality checkers reliable? adversarial meta-evaluation of factuality in summarization.
\newblock In \emph{EMNLP}.

\bibitem[{Fabbri et~al.(2021)Fabbri, Kry{\'s}ci{\'n}ski, McCann, Xiong, Socher, and Radev}]{fabbri2021summeval}
Alexander~R Fabbri, Wojciech Kry{\'s}ci{\'n}ski, Bryan McCann, Caiming Xiong, Richard Socher, and Dragomir Radev. 2021.
\newblock Summeval: Re-evaluating summarization evaluation.
\newblock \emph{Transactions of the Association for Computational Linguistics}, 9:391--409.

\bibitem[{Fabbri et~al.(2022)Fabbri, Wu, Liu, and Xiong}]{fabbri2022qafacteval}
Alexander~Richard Fabbri, Chien-Sheng Wu, Wenhao Liu, and Caiming Xiong. 2022.
\newblock {QAF}act{E}val: Improved qa-based factual consistency evaluation for summarization.
\newblock In \emph{NAACL}.

\bibitem[{Feng et~al.(2023)Feng, Balachandran, Bai, and Tsvetkov}]{feng2023factkb}
Shangbin Feng, Vidhisha Balachandran, Yuyang Bai, and Yulia Tsvetkov. 2023.
\newblock Fact{KB}: Generalizable factuality evaluation using language models enhanced with factual knowledge.
\newblock In \emph{EMNLP}.

\bibitem[{Fu et~al.(2023)Fu, Ng, Jiang, and Liu}]{fu2023gptscore}
Jinlan Fu, See-Kiong Ng, Zhengbao Jiang, and Pengfei Liu. 2023.
\newblock {GPT}score: Evaluate as you desire.
\newblock \emph{arXiv preprint arXiv:2302.04166}.

\bibitem[{Gao et~al.(2023)Gao, Ruan, Sun, Yin, Yang, and Wan}]{gao2023human}
Mingqi Gao, Jie Ruan, Renliang Sun, Xunjian Yin, Shiping Yang, and Xiaojun Wan. 2023.
\newblock Human-like summarization evaluation with chatgpt.
\newblock \emph{arXiv preprint arXiv:2304.02554}.

\bibitem[{Gao and Wan(2022)}]{gao2022dialsummeval}
Mingqi Gao and Xiaojun Wan. 2022.
\newblock Dialsummeval: Revisiting summarization evaluation for dialogues.
\newblock In \emph{NAACL}.

\bibitem[{Gekhman et~al.()Gekhman, Herzig, Aharoni, Elkind, and Szpektor}]{gekhman2023trueteacher}
Zorik Gekhman, Jonathan Herzig, Roee Aharoni, Chen Elkind, and Idan Szpektor.
\newblock True{T}eacher: Learning factual consistency evaluation with large language models.
\newblock In \emph{EMNLP}.

\bibitem[{Glover et~al.(2022)Glover, Fancellu, Jagannathan, Gormley, and Schaaf}]{glover2022revisiting}
John Glover, Federico Fancellu, Vasudevan Jagannathan, Matthew~R Gormley, and Thomas Schaaf. 2022.
\newblock Revisiting text decomposition methods for nli-based factuality scoring of summaries.
\newblock In \emph{GEM}.

\bibitem[{Goyal and Durrett(2020)}]{goyal2020evaluating}
Tanya Goyal and Greg Durrett. 2020.
\newblock Evaluating factuality in generation with dependency-level entailment.
\newblock In \emph{EMNLP}.

\bibitem[{Gupta and Gupta(2019)}]{gupta2019abstractive}
Som Gupta and Sanjai~Kumar Gupta. 2019.
\newblock Abstractive summarization: An overview of the state of the art.
\newblock \emph{Expert Systems with Applications}, 121:49--65.

\bibitem[{Hermann et~al.(2015)Hermann, Kocisky, Grefenstette, Espeholt, Kay, Suleyman, and Blunsom}]{hermann2015teaching}
Karl~Moritz Hermann, Tomas Kocisky, Edward Grefenstette, Lasse Espeholt, Will Kay, Mustafa Suleyman, and Phil Blunsom. 2015.
\newblock Teaching machines to read and comprehend.
\newblock In \emph{NeurIPS}.

\bibitem[{Honovich et~al.(2022)Honovich, Aharoni, Herzig, Taitelbaum, Kukliansy, Cohen, Scialom, Szpektor, Hassidim, and Matias}]{honovich2022true}
Or~Honovich, Roee Aharoni, Jonathan Herzig, Hagai Taitelbaum, Doron Kukliansy, Vered Cohen, Thomas Scialom, Idan Szpektor, Avinatan Hassidim, and Yossi Matias. 2022.
\newblock {TRUE}: Re-evaluating factual consistency evaluation.
\newblock In \emph{NAACL}.

\bibitem[{Ji et~al.(2023)Ji, Lee, Frieske, Yu, Su, Xu, Ishii, Bang, Madotto, and Fung}]{ji2023survey}
Ziwei Ji, Nayeon Lee, Rita Frieske, Tiezheng Yu, Dan Su, Yan Xu, Etsuko Ishii, Ye~Jin Bang, Andrea Madotto, and Pascale Fung. 2023.
\newblock Survey of hallucination in natural language generation.
\newblock \emph{ACM Computing Surveys}, 55(12):1--38.

\bibitem[{Jiang et~al.(2024)Jiang, Sablayrolles, Roux, Mensch, Savary, Bamford, Chaplot, Casas, Hanna, Bressand et~al.}]{jiang2024mixtral}
Albert~Q Jiang, Alexandre Sablayrolles, Antoine Roux, Arthur Mensch, Blanche Savary, Chris Bamford, Devendra~Singh Chaplot, Diego de~las Casas, Emma~Bou Hanna, Florian Bressand, et~al. 2024.
\newblock Mixtral of experts.
\newblock \emph{arXiv preprint arXiv:2401.04088}.

\bibitem[{Kry{\'s}ci{\'n}ski et~al.(2020)Kry{\'s}ci{\'n}ski, McCann, Xiong, and Socher}]{kryscinski2020evaluating}
Wojciech Kry{\'s}ci{\'n}ski, Bryan McCann, Caiming Xiong, and Richard Socher. 2020.
\newblock Evaluating the factual consistency of abstractive text summarization.
\newblock In \emph{EMNLP}.

\bibitem[{Laban et~al.(2023)Laban, Kry{\'s}ci{\'n}ski, Agarwal, Fabbri, Xiong, Joty, and Wu}]{laban2023summedits}
Philippe Laban, Wojciech Kry{\'s}ci{\'n}ski, Divyansh Agarwal, Alexander~Richard Fabbri, Caiming Xiong, Shafiq Joty, and Chien-Sheng Wu. 2023.
\newblock Summ{E}dits: Measuring llm ability at factual reasoning through the lens of summarization.
\newblock In \emph{EMNLP}.

\bibitem[{Laban et~al.(2022)Laban, Schnabel, Bennett, and Hearst}]{laban2022summac}
Philippe Laban, Tobias Schnabel, Paul~N Bennett, and Marti~A Hearst. 2022.
\newblock Summa{C}: Re-visiting nli-based models for inconsistency detection in summarization.
\newblock \emph{Transactions of the Association for Computational Linguistics}, 10:163--177.

\bibitem[{Lewis et~al.(2020)Lewis, Liu, Goyal, Ghazvininejad, Mohamed, Levy, Stoyanov, and Zettlemoyer}]{lewis2020bart}
Mike Lewis, Yinhan Liu, Naman Goyal, Marjan Ghazvininejad, Abdelrahman Mohamed, Omer Levy, Veselin Stoyanov, and Luke Zettlemoyer. 2020.
\newblock {BART}: Denoising sequence-to-sequence pre-training for natural language generation, translation, and comprehension.
\newblock In \emph{ACL}.

\bibitem[{Lin(2004)}]{lin2004rouge}
Chin-Yew Lin. 2004.
\newblock {ROUGE}: A package for automatic evaluation of summaries.
\newblock In \emph{Text Summarization Branches Out}.

\bibitem[{Lin and Chen(2023)}]{lin2023llm}
Yen-Ting Lin and Yun-Nung Chen. 2023.
\newblock Llm-eval: Unified multi-dimensional automatic evaluation for open-domain conversations with large language models.
\newblock \emph{arXiv preprint arXiv:2305.13711}.

\bibitem[{Liu et~al.(2023)Liu, Iter, Xu, Wang, Xu, and Zhu}]{liu2023gpteval}
Yang Liu, Dan Iter, Yichong Xu, Shuohang Wang, Ruochen Xu, and Chenguang Zhu. 2023.
\newblock {G}-{E}val: {NLG} evaluation using gpt-4 with better human alignment.
\newblock \emph{arXiv preprint arXiv:2303.16634}.

\bibitem[{Maynez et~al.(2020)Maynez, Narayan, Bohnet, and McDonald}]{maynez2020faithfulness}
Joshua Maynez, Shashi Narayan, Bernd Bohnet, and Ryan McDonald. 2020.
\newblock On faithfulness and factuality in abstractive summarization.
\newblock In \emph{ACL}.

\bibitem[{Min et~al.(2023)Min, Krishna, Lyu, Lewis, Yih, Koh, Iyyer, Zettlemoyer, and Hajishirzi}]{min2023factscore}
Sewon Min, Kalpesh Krishna, Xinxi Lyu, Mike Lewis, Wen-tau Yih, Pang Koh, Mohit Iyyer, Luke Zettlemoyer, and Hannaneh Hajishirzi. 2023.
\newblock {FA}ct{S}core: Fine-grained atomic evaluation of factual precision in long form text generation.
\newblock In \emph{EMNLP}.

\bibitem[{Narayan et~al.(2018)Narayan, Cohen, and Lapata}]{Narayan2018DontGM}
Shashi Narayan, Shay~B. Cohen, and Mirella Lapata. 2018.
\newblock Don't give me the details, just the summary! topic-aware convolutional neural networks for extreme summarization.
\newblock \emph{ArXiv}, abs/1808.08745.

\bibitem[{Pagnoni et~al.(2021)Pagnoni, Balachandran, and Tsvetkov}]{pagnoni2021understanding}
Artidoro Pagnoni, Vidhisha Balachandran, and Yulia Tsvetkov. 2021.
\newblock Understanding factuality in abstractive summarization with frank: A benchmark for factuality metrics.
\newblock In \emph{NAACL}.

\bibitem[{Papineni et~al.(2002)Papineni, Roukos, Ward, and Zhu}]{papineni2002bleu}
Kishore Papineni, Salim Roukos, Todd Ward, and Wei-Jing Zhu. 2002.
\newblock {BLEU}: A method for automatic evaluation of machine translation.
\newblock In \emph{ACL}.

\bibitem[{Saito et~al.(2023)Saito, Wachi, Wataoka, and Akimoto}]{saito2023verbosity}
Keita Saito, Akifumi Wachi, Koki Wataoka, and Youhei Akimoto. 2023.
\newblock Verbosity bias in preference labeling by large language models.
\newblock \emph{arXiv preprint arXiv:2310.10076}.

\bibitem[{Scialom et~al.(2021)Scialom, Dray, Gallinari, Lamprier, Piwowarski, Staiano, and Wang}]{scialom2021questeval}
Thomas Scialom, Paul-Alexis Dray, Patrick Gallinari, Sylvain Lamprier, Benjamin Piwowarski, Jacopo Staiano, and Alex Wang. 2021.
\newblock Quest{E}val: Summarization asks for fact-based evaluation.
\newblock In \emph{EMNLP}.

\bibitem[{Shen et~al.(2023)Shen, Cheng, Nguyen, You, and Bing}]{shen2023large}
Chenhui Shen, Liying Cheng, Xuan-Phi Nguyen, Yang You, and Lidong Bing. 2023.
\newblock Large language models are not yet human-level evaluators for abstractive summarization.
\newblock In \emph{EMNLP}.

\bibitem[{Shen and Wan(2023)}]{shen2023opinsummeval}
Yuchen Shen and Xiaojun Wan. 2023.
\newblock Opinsummeval: Revisiting automated evaluation for opinion summarization.
\newblock \emph{arXiv preprint arXiv:2310.18122}.

\bibitem[{Shi et~al.(2023)Shi, Xu, Ding, Pang, Liu, Luo, Peng, Lu, Yang, Hu et~al.}]{shi2023llm}
Xiaoming Shi, Jie Xu, Jinru Ding, Jiali Pang, Sichen Liu, Shuqing Luo, Xingwei Peng, Lu~Lu, Haihong Yang, Mingtao Hu, et~al. 2023.
\newblock Llm-mini-cex: Automatic evaluation of large language model for diagnostic conversation.
\newblock \emph{arXiv preprint arXiv:2308.07635}.

\bibitem[{Song et~al.(2023)Song, Shalyminov, Su, Siffi, Yao, and Mansour}]{song2023enhancing}
Hwanjun Song, Igor Shalyminov, Hang Su, Singh Siffi, Kaisheng Yao, and Saab Mansour. 2023.
\newblock Enhancing abstractiveness of summarization models through calibrated distillation.
\newblock In \emph{EMNLP}.

\bibitem[{Touvron et~al.(2023)Touvron, Martin, Stone, Albert, Almahairi, Babaei, Bashlykov, Batra, Bhargava, Bhosale et~al.}]{touvron2023llama}
Hugo Touvron, Louis Martin, Kevin Stone, Peter Albert, Amjad Almahairi, Yasmine Babaei, Nikolay Bashlykov, Soumya Batra, Prajjwal Bhargava, Shruti Bhosale, et~al. 2023.
\newblock Llama 2: Open foundation and fine-tuned chat models.
\newblock \emph{arXiv preprint arXiv:2307.09288}.

\bibitem[{Utama et~al.(2022)Utama, Bambrick, Moosavi, and Gurevych}]{utama2022falsesum}
PA~Utama, J~Bambrick, NS~Moosavi, and I~Gurevych. 2022.
\newblock Falsesum: generating document-level nli examples for recognizing factual inconsistency in summarization.
\newblock In \emph{NAACL}.

\bibitem[{Wang et~al.(2020)Wang, Cho, and Lewis}]{wang2020asking}
Alex Wang, Kyunghyun Cho, and Mike Lewis. 2020.
\newblock Asking and answering questions to evaluate the factual consistency of summaries.
\newblock In \emph{ACL}.

\bibitem[{Wang et~al.(2023)Wang, Liang, Meng, Shi, Li, Xu, Qu, and Zhou}]{wang2023chatgpt}
Jiaan Wang, Yunlong Liang, Fandong Meng, Haoxiang Shi, Zhixu Li, Jinan Xu, Jianfeng Qu, and Jie Zhou. 2023.
\newblock Is chatgpt a good nlg evaluator? a preliminary study.
\newblock \emph{arXiv preprint arXiv:2303.04048}.

\bibitem[{Wei et~al.(2022)Wei, Wang, Schuurmans, Bosma, Xia, Chi, Le, Zhou et~al.}]{wei2022chain}
Jason Wei, Xuezhi Wang, Dale Schuurmans, Maarten Bosma, Fei Xia, Ed~Chi, Quoc~V Le, Denny Zhou, et~al. 2022.
\newblock Chain-of-thought prompting elicits reasoning in large language models.
\newblock In \emph{NeurIPS}.

\bibitem[{Yu et~al.(2023)Yu, He, Wu, Dai, and Chen}]{yu2023towards}
Zihan Yu, Liang He, Zhen Wu, Xinyu Dai, and Jiajun Chen. 2023.
\newblock Towards better chain-of-thought prompting strategies: A survey.
\newblock \emph{arXiv preprint arXiv:2310.04959}.

\bibitem[{Yuan et~al.(2021)Yuan, Neubig, and Liu}]{yuan2021bartscore}
Weizhe Yuan, Graham Neubig, and Pengfei Liu. 2021.
\newblock {BARTS}core: Evaluating generated text as text generation.
\newblock In \emph{NeurIPS}.

\bibitem[{Zhang et~al.(2023)Zhang, Dong, Li, Zhang, Sun, Wang, Li, Hu, Zhang, Wu et~al.}]{zhang2023instruction}
Shengyu Zhang, Linfeng Dong, Xiaoya Li, Sen Zhang, Xiaofei Sun, Shuhe Wang, Jiwei Li, Runyi Hu, Tianwei Zhang, Fei Wu, et~al. 2023.
\newblock Instruction tuning for large language models: A survey.
\newblock \emph{arXiv preprint arXiv:2308.10792}.

\bibitem[{Zhang et~al.(2019)Zhang, Kishore, Wu, Weinberger, and Artzi}]{zhang2019bertscore}
Tianyi Zhang, Varsha Kishore, Felix Wu, Kilian~Q Weinberger, and Yoav Artzi. 2019.
\newblock {BERTS}core: Evaluating text generation with bert.
\newblock In \emph{ICLR}.

\bibitem[{Zhao et~al.(2019)Zhao, Peyrard, Liu, Gao, Meyer, and Eger}]{zhao2019moverscore}
Wei Zhao, Maxime Peyrard, Fei Liu, Yang Gao, Christian~M Meyer, and Steffen Eger. 2019.
\newblock Mover{S}core: Text generation evaluating with contextualized embeddings and earth mover distance.
\newblock In \emph{EMNLP}.

\bibitem[{Zhong et~al.(2022)Zhong, Liu, Yin, Mao, Jiao, Liu, Zhu, Ji, and Han}]{zhong2022towards}
Ming Zhong, Yang Liu, Da~Yin, Yuning Mao, Yizhu Jiao, Pengfei Liu, Chenguang Zhu, Heng Ji, and Jiawei Han. 2022.
\newblock Towards a unified multi-dimensional evaluator for text generation.
\newblock In \emph{EMNLP}.

\end{thebibliography}


\appendix

\begin{appendix}
\clearpage
\begin{table*}[t]
\begin{center}
\footnotesize
\begin{tabular}{|L{0.8cm} L{2.7cm} |L{5.4cm} L{5.4cm} |}\toprule
 & Category & Description & Example \\ \midrule
OutE  & Out of context Error    & The statement contains information not present in the source article. & \emph{{\color{red}China} has already started clinical trials of the COVID-19 vaccine.}\\\midrule
EntE  & Entity Error            & The primary arguments (or their attributes) of the predicate are wrong. & \emph{The {\color{red}COVID-19 vaccine} was approved by the FDA in 2019.} \\\midrule
PredE & Predicate Error     & The predicate in the summary statement is inconsistent with the source article. & \emph{The Ebola vaccine {\color{red}was rejected} by the FDA in 2019.} \\\midrule
CirE  & Circumstance Error  & The additional information (like location or time) specifying the circumstance around a predicate is wrong. & \emph{The first vaccine for Ebola was approved by the FDA in {\color{red}2014}.} \\\midrule
GramE\!\!\! & Grammatical Error   &  The grammar of the sentence is so wrong that it becomes meaningless. & \emph{The Ebola vaccine {\color{red}accepted have already started}.} \\\midrule
LinkE & Discourse Link Error & Error in how multiple statements are linked together in the discourse (for example tempral ordering/causal link). & \emph{To produce the vaccine, scientists have to show successful human trials, {\color{red}then} sequence the DNA of the virus.} \\\midrule
CorefE\!\!\! & Coreference Error   & A pronoun/reference with wrong or non-existing antecedent. & \emph{The first vaccine for Ebola was approved in 2019. {\color{red}They} say a vaccine for COVID-19 is unlikely to be ready this year.} \\\bottomrule
\end{tabular}
\end{center}
\vspace*{-0.4cm}
\caption{\textbf{Typology of factual errors copied from \cite{pagnoni2021understanding}.} Original text for the examples: \emph{The first vaccine for Ebola was approved by the FDA in 2019 in the US, five years after the initial outbreak in 2014. To produce the vaccine, scientists had to sequence the DNA of Ebola, then identify possible vaccines, and finally show successful clinical trials. Scientists say a vaccine for COVID-19 is unlikely to be ready this year, although clinical trials have already started.}} 
\label{table:error_type}
\vspace*{-0.3cm}
\end{table*}

\pagebreak

\section{Factuality Error Type}
\label{sec:category_example}

Following the work \cite{fu-frank-2023-seti}, we use seven categories to define factuality error types, namely "out of context", "predicate," "entity," "circumstance," "coreference," "discourse link," and "grammatical". Table \ref{table:error_type} provides the detailed description and example of the error categories.

\begin{figure}[t!]
\begin{center}
\includegraphics[width=7.7cm]{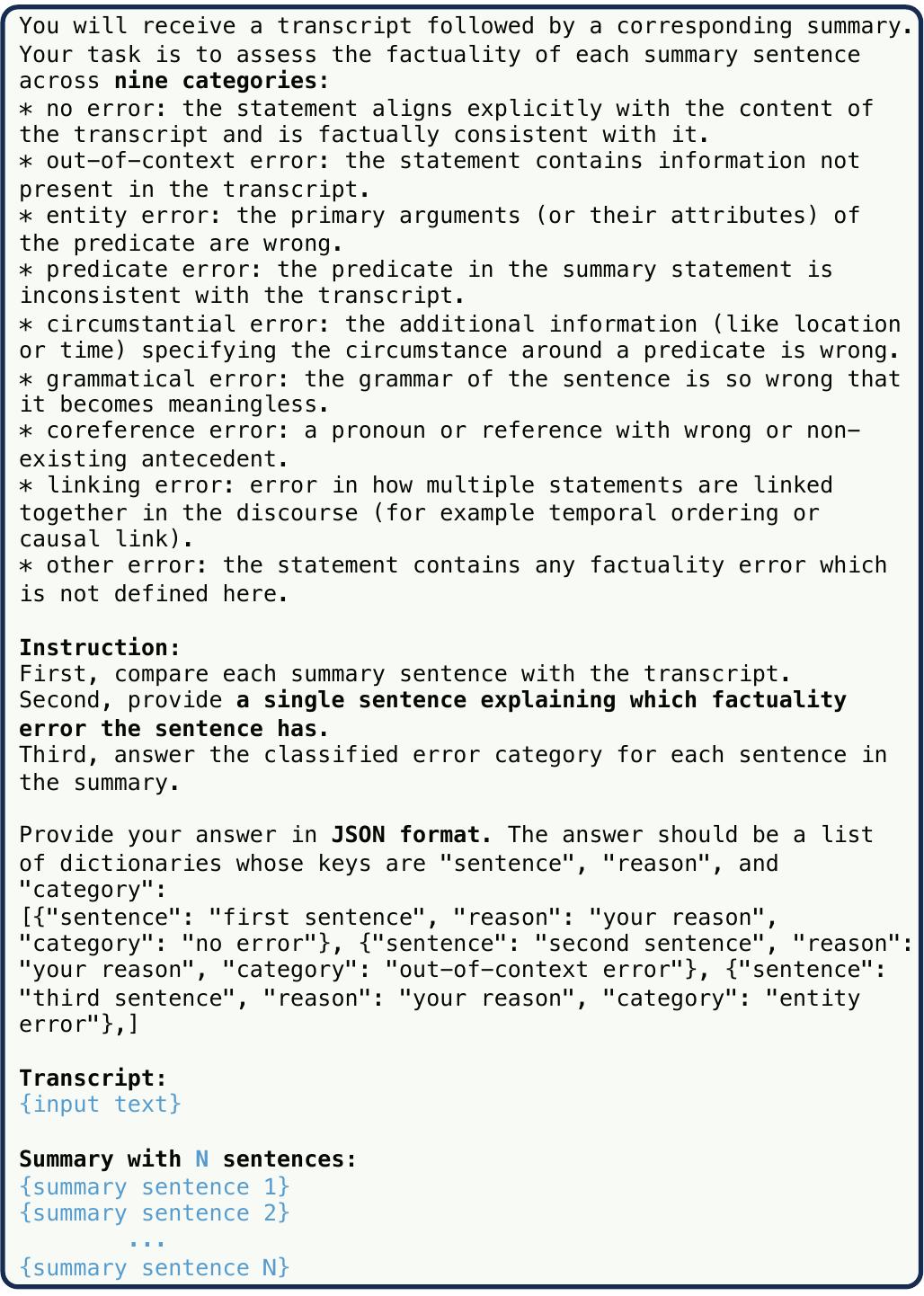}
\end{center}
\vspace*{-0.47cm}
\caption{\textbf{Prompt for fact checking:} the prompt is tailored for a categorization task, utilizing an instruction format with a structured reasoning step. For every summary sentence, the output is a dictionary that provides the category (one of the error types) along with a concise sentence of reasoning.}
\vspace*{-0.45cm}
\label{fig:best_prompt_factuality}
\end{figure}

\section{Main Prompt}
\label{sec:main_prompt}

We employ LLMs as a tool to conduct fact checking and keyfact alignment tasks. Specifically, we design two prompts tailored for the two tasks, as shown in Figures \ref{fig:best_prompt_factuality}-\ref{fig:best_prompt_keyfact}.

\section{Keyfact Extraction}
\label{sec:deatail_keyfact_extraction}

The list of key facts is crucial for evaluating completeness and conciseness. Ideally, they should be generated by humans, as the key facts in text summarization heavily depend on what information humans prioritize in various domains. 
For instance, in a medical scenario, keyfacts should encompass all medical symptoms and the doctor's recommended treatment, while in a sales call, the customer's issue and action should be prioritized as the key facts.

\begin{figure}[t!]
\begin{center}
\includegraphics[width=7.7cm]{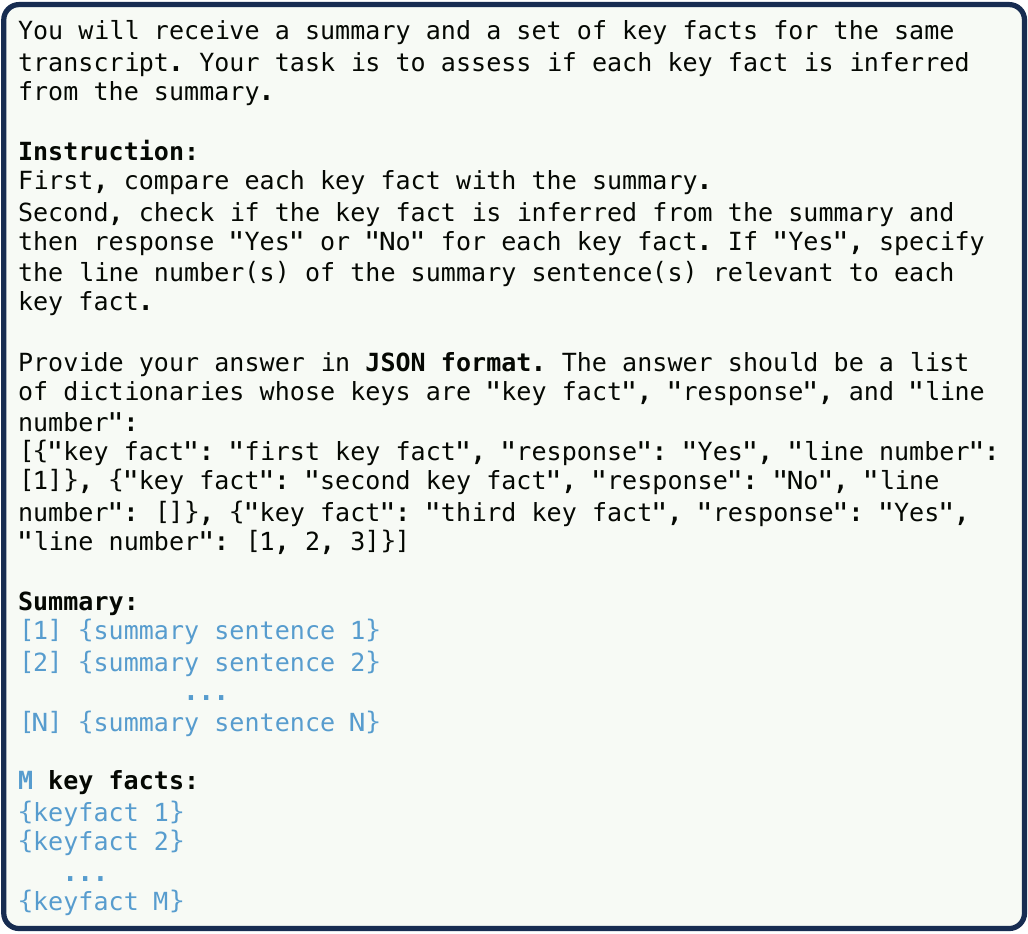}
\end{center}
\vspace*{-0.45cm}
\caption{\textbf{Prompt for keyfact alignment:} the prompt is tailored for keyfact matching, employing a simple instruction format. For every keyfact, the output is a dictionary that provides a binary labeling of "Yes" (align) or "No" (not align) and all aligned sentence IDs.}
\vspace*{-0.15cm}
\label{fig:best_prompt_keyfact}
\end{figure}

\begin{figure}[t!]
\begin{center}
\includegraphics[width=7.7cm]{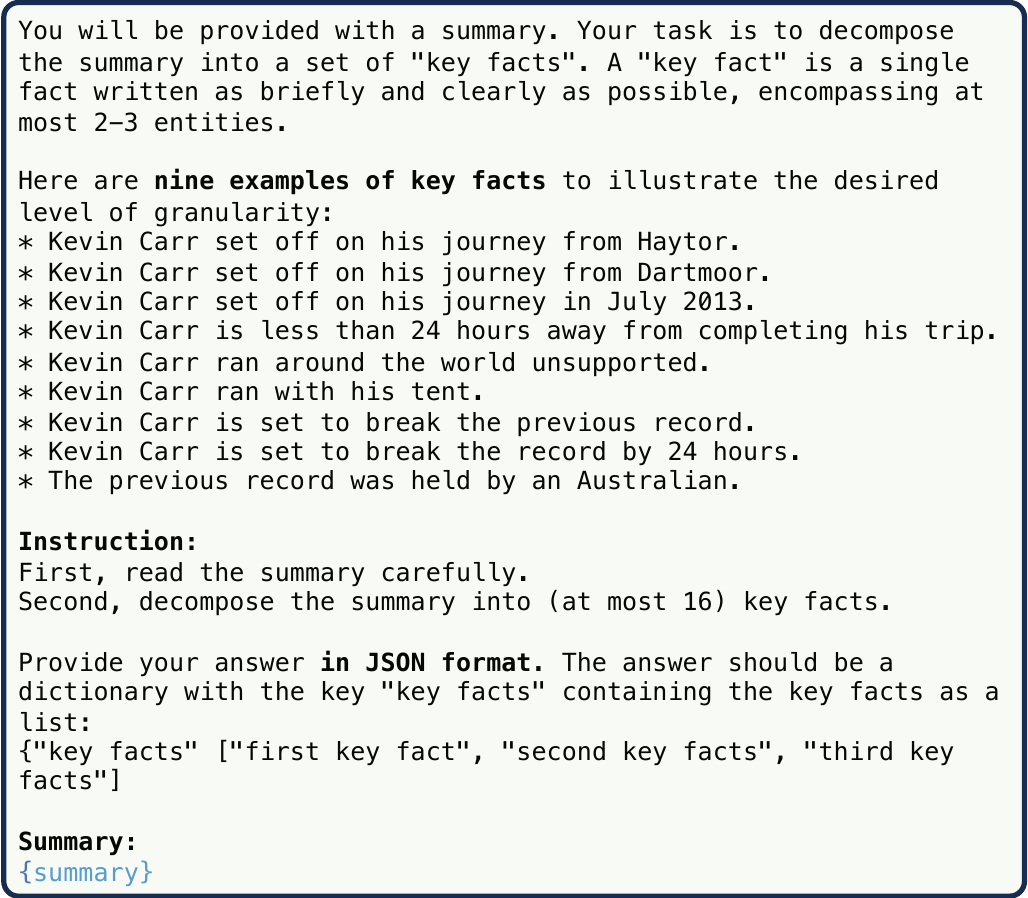}
\end{center}
\vspace*{-0.35cm}
\caption{\textbf{Prompt for keyfact extraction:} The prompt is tailored for extracting keyfacts, utilizing an instruction format with few-shot examples. Given a reference summary, the output is a dictionary that provides a list of keyfacts. We adhere to the REALSum's annotation guideline, which limits the number of key facts to 16.}
\vspace*{-0.3cm}
\label{fig:keyfact_extract}
\end{figure}

We also automatically extract feasible keyfacts from the reference summary, similar to the approach taken by REALSumm in annotating key facts with human annotators. However, our process is fully automatic, utilizing LLMs instead of human's manual efforts. We designed the prompt tailored for extracting keyfacts in Figure \ref{fig:keyfact_extract}. This prompt generates up to 16 key facts from the reference summaries by providing few-shot examples to ensure that the granularity of extracted key facts aligns with our requirements.

\section{Measurement}
\label{sec:measurement}
We utilize several measurement to compute the agreement with human judgements at the three different levels.

For sentence-level assessment, we utilize human binary annotations indicating the presence of factuality errors for each sentence, denoted as "0" for no error and "1" for error. Similarly, the LLM returns the binary decision of "0" (No) and "1" (Yes) by the fact checking prompt per sentence, as shown in Figure \ref{fig:best_prompt_factuality}. Then, the \emph{balanced accuracy (bACC)} is computed by:
\begin{equation}
{\rm bACC} = ({\rm sensitivity} + {\rm specificity}) / 2,
\end{equation}
where {sensitivity} refers to the true positive rate, which measures the proportion of correct predictions by LLMs out of all positive predictions. On the other hand, {specificity} is the true negative rate, measuring the proportion of correct predictions out of all negative predictions.

For the summary-level assessment, let $D=\{d_1, \dots, d_k\}$ be the set of input documents and $S=\{s_1, \dots, s_k\}$ be the set of summaries corresponding to the document set. Supposing that $F_{gt}$ and $F_{pred}$ are the functions that returns the percentage score of a specific evaluation dimensions based on human and predicted labels, respectively ($F$ can be any scoring function in Eq.\,\eqref{eq:faithfulness_score}-\eqref{eq:other_scores}). Then, the \emph{summary-level} correlation is calculated as follows:
\begin{equation}
\begin{gathered}
{\rm Corr}\Big(\big[F_{gt}(d_1, s_1), \dots, F_{gt}(d_k, s_k)\big],\\
\big[F_{pred}(d_1, s_1), \dots, F_{pred}(d_k, s_k)\big]\Big),
\end{gathered}
\end{equation}
where {Corr} is one of the Pearson and Spearman correlation measure. 
The measurement reveals the agreement between the percentage scores generated by automated evaluators and human judgments.

\begin{figure}[t!]
\begin{center}
\includegraphics[width=7.7cm]{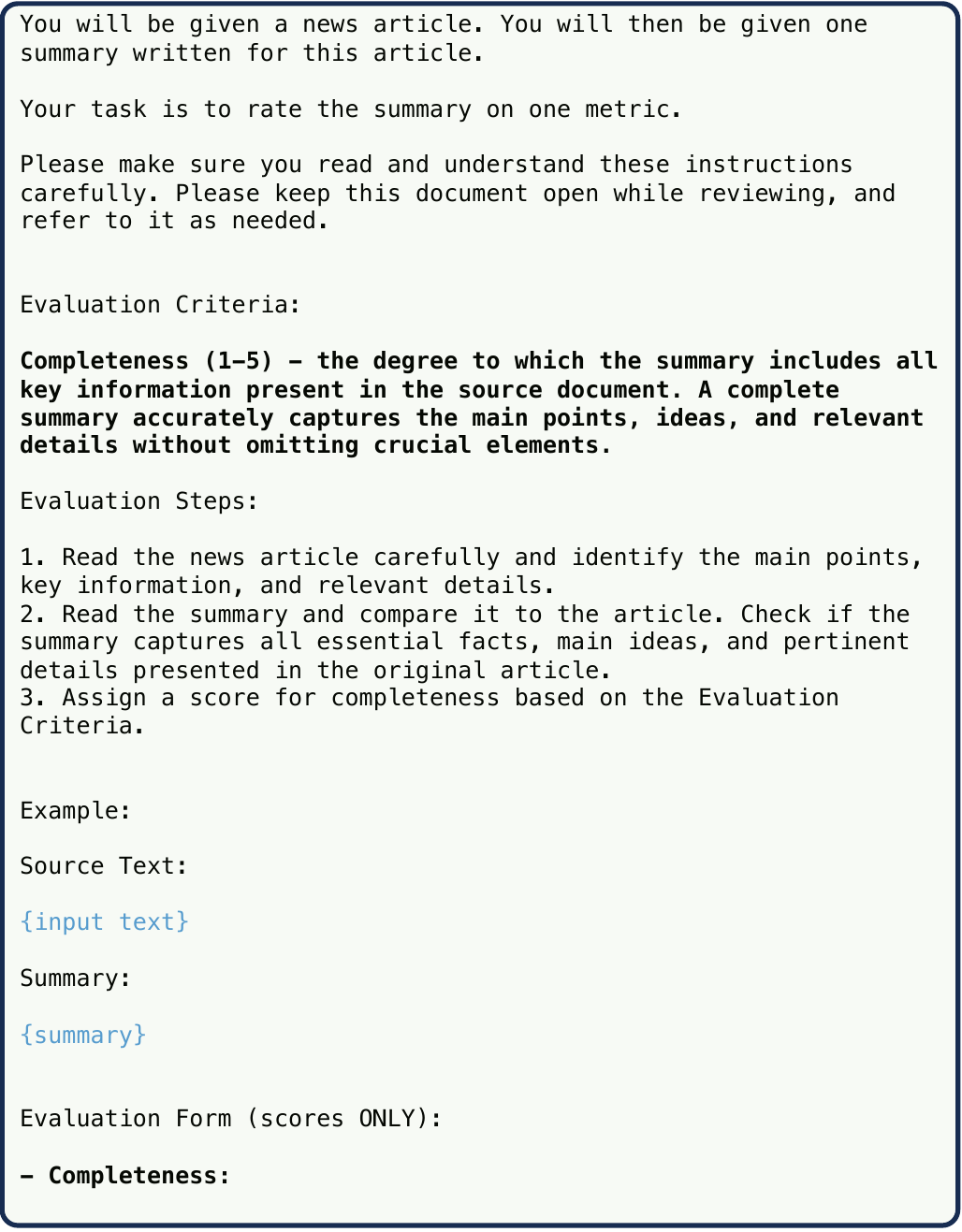}
\end{center}
\vspace*{-0.42cm}
\caption{G-Eval for completeness \textbf{without} keyfacts.}
\vspace*{-0.3cm}
\label{fig:geval-completeness-wo-keyfacts}
\end{figure}

\begin{figure}[t!]
\begin{center}
\includegraphics[width=7.7cm]{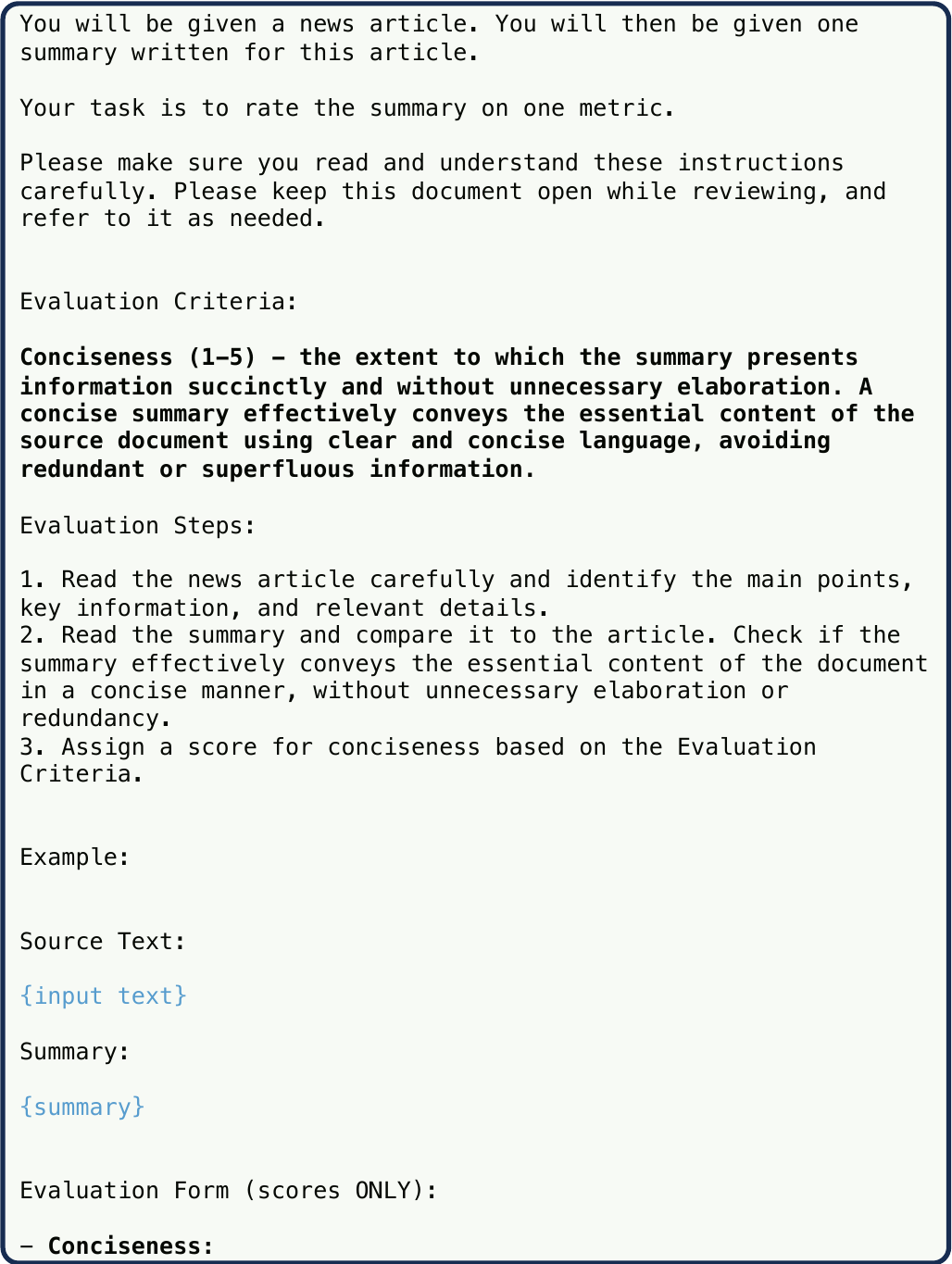}
\end{center}
\vspace*{-0.4cm}
\caption{G-Eval for conciseness \textbf{without} keyfacts.}
\vspace*{-0.3cm}
\label{fig:geval-completeness-w-keyfacts}
\end{figure}

\begin{table*}[t!]
\begin{center}
\includegraphics[width=16.1cm]{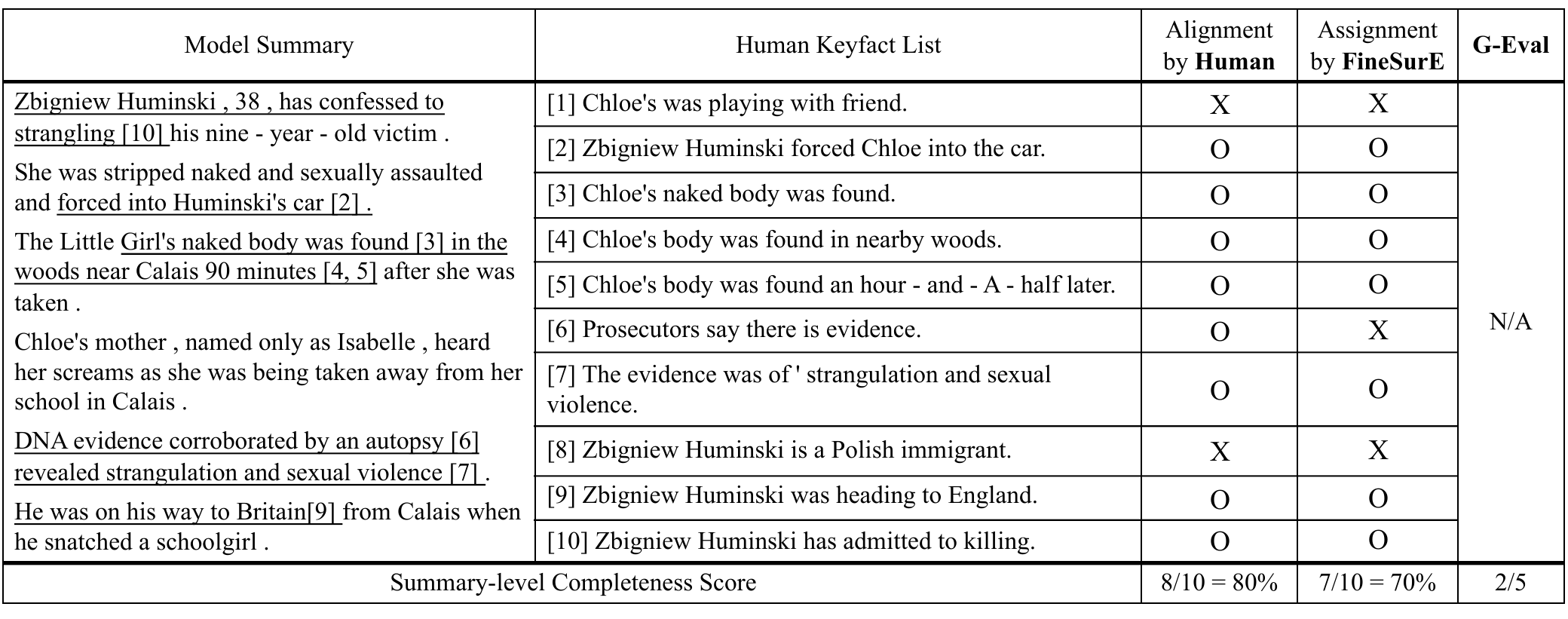}
\end{center}
\vspace*{-0.55cm}
\caption{\textbf{Qualitative analysis of completeness evaluation:} an example showing the limitations of the existing Likert-scale based evaluation using LLM (G-Eval). In "Model Summary" column, all the statements aligned with keyfacts are \underline{underlined} with the keyfact number, e.g., [1]. The last row indicates the summary-level completeness scores from human, our fine-grained \algname{}, and likert-scale based G-Eval.}
\vspace*{-0.3cm}
\label{table:qualitative_completeness}
\end{table*}

For system-level assessment, we consolidate the percentage scores across all input documents , determining the average percentage score for each summarization model. 
Let $\mathbb{F}_{m}=\{F_{m}(d_1, s_1), \dots, F_{m}(d_k, s_k)\}$ represents the percentage scores derived from the labels assigned by a summarization system $m$. Then, we make a list of the average percentage score for all summarization systems, $[\bar{\mathbb{F}_{m_1}}, \bar{\mathbb{F}_{m_2}}, \dots]$ and compute their ranking by using the ${\rm Rank}$ function, returning the list $[{\rm rank}_{m_1}, {\rm rank}_{m_2}, \dots]$, where ${\rm rank}_m$ is the ranking of the model $m$. Given a list of ground-truth rankings $[{\rm rank}^{*}_{m_1}, {rank}^{*}_{m_2}, \dots]$ using the human scores, we compute the \emph{rank correlation} by:
\begin{equation}
\begin{gathered}
{\rm Spearman}\Big(\big[{\rm rank}_{m_1}, {\rm rank}_{m_2}, \dots\big], \\
\big[{\rm rank}^{*}_{m_1}, {rank}^{*}_{m_2}, \dots \big]\Big).
\end{gathered}
\end{equation}


\vspace*{-0.1cm}
\section{Qualitative Analysis for Completeness}
\label{sec:qualitatitve_completeness}
\vspace*{-0.1cm}

\begin{figure}[b!]
\begin{center}
\includegraphics[width=7.7cm]{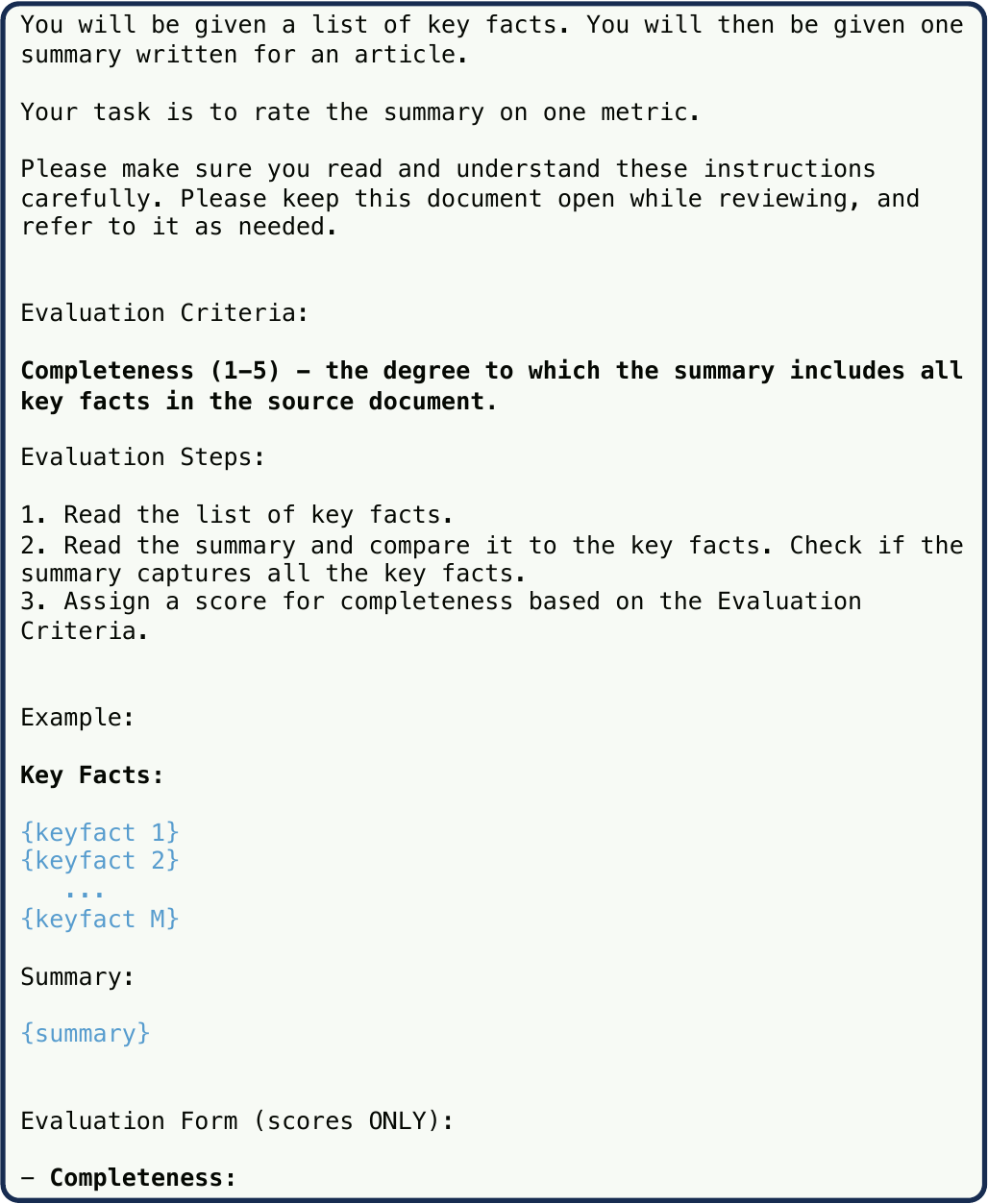}
\end{center}
\vspace*{-0.42cm}
\caption{G-Eval for completeness \textbf{with} keyfacts.}
\vspace*{-0.3cm}
\label{fig:geval-conciseness-wo-keyfacts}
\end{figure}

\begin{figure}[b!]
\begin{center}
\includegraphics[width=7.7cm]{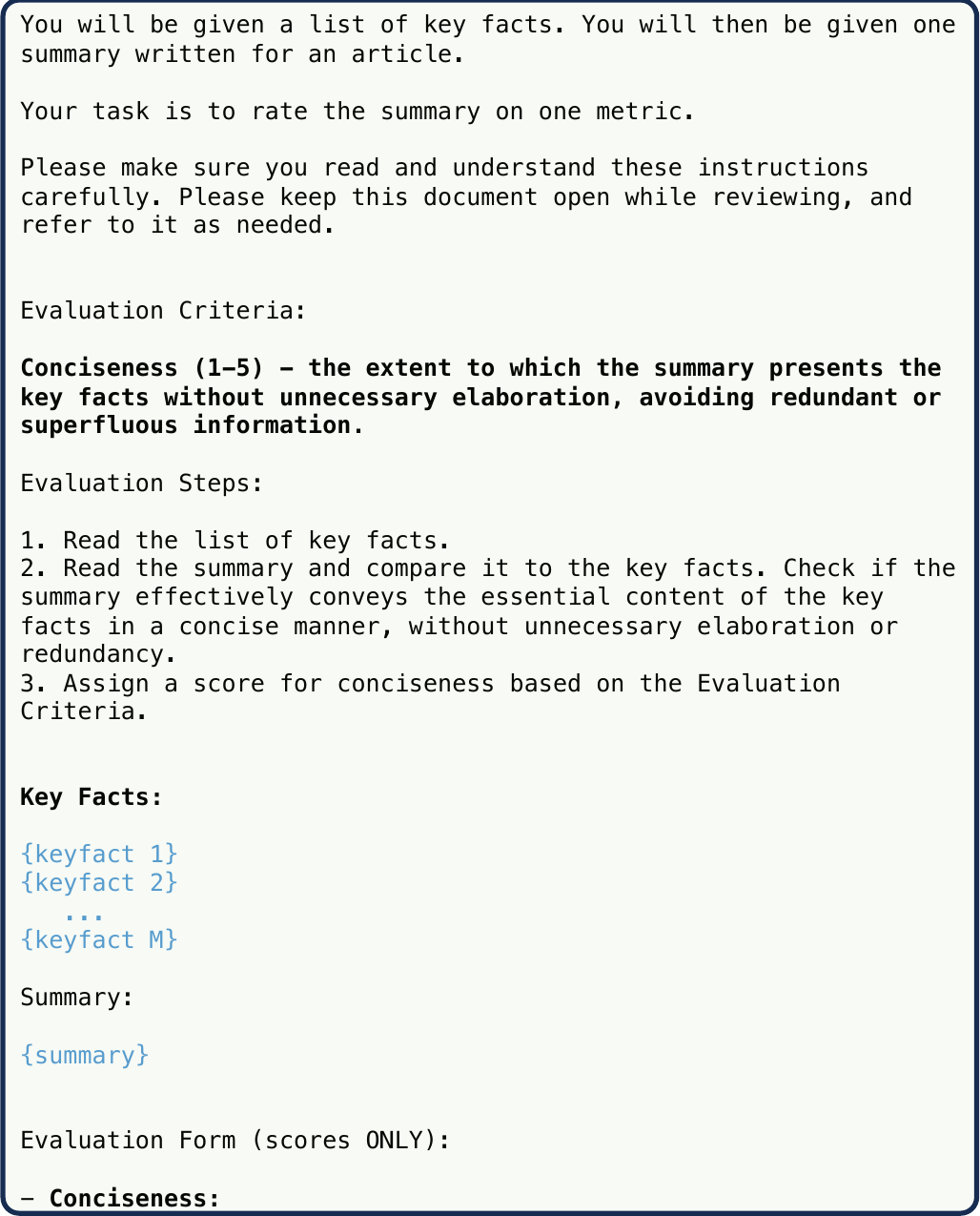}
\end{center}
\vspace*{-0.42cm}
\caption{G-Eval for conciseness \textbf{with} keyfacts.}
\label{fig:geval-conciseness-w-keyfacts}
\end{figure}

\begin{figure}[b!]
\begin{center}
\includegraphics[width=7.7cm]{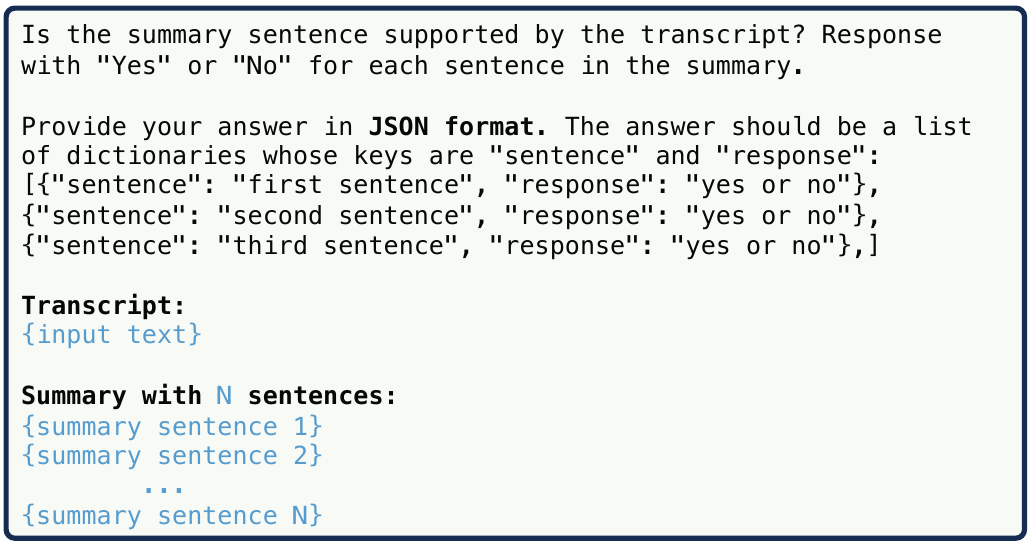}
\end{center}
\vspace*{-0.42cm}
\caption{Basic prompt for faithfulness evaluation.}
\vspace*{-0.3cm}
\label{fig:faith_basic}
\end{figure}

Table \ref{table:qualitative_completeness} shows an example showing the limitation of the existing Likert-scale based evaluation method, such as G-Eval. Although the model summary includes eight out of ten human keyfacts, G-Eval output a very low completeness score, i.e., two out of five. However, the proposed \algname{} exhibits the  percentage score similar to that of human judgement. This is because our framework conducts a human-like keyfact alignment to determine the completeness score, which is more fine-grained than relying solely on Likert-scale judgments. Therefore, a fine-grained evaluation framework has great potential to enhance the quality of automated evaluation using LLMs.

\begin{table*}[t]
\begin{center}
\footnotesize
\begin{tabular}{|L{2.8cm} |X{1.4cm} X{1.4cm} X{1.5cm} |X{1.4cm} X{1.4cm} X{1.5cm}|}\toprule
Evaluation Dim. & \multicolumn{3}{c|}{(a) Completeness} & \multicolumn{3}{c|}{(b) Conciseness}  \\\midrule
Method &  \multicolumn{2}{c}{Summary-level} & \!\!\!System-level\!\!\! & \multicolumn{2}{c}{Summary-level} & \!\!\!System-level\!\!\!  \\ 
 & \!\!\!Pearson\,($\uparrow$)\!\!\! & \!\!\!Spearman\,($\uparrow$)\!\!\! & Rank\,($\uparrow$) & \!\!\!Pearson\,($\uparrow$)\!\!\! & \!\!\!Spearman\,($\uparrow$)\!\!\! & Rank\,($\uparrow$) \\ \midrule
G-Eval\,(GPT4) & \!\!\!0.314\!\!\! & \!\!\!0.295\!\!\! & \!\!\!0.908\!\!\! & \!\!\!0.314\!\!\! & \!\!\!0.277\!\!\! & \!\!\!0.582\!\!\! \\
~~~+ Adjusted Criteria & \!\!\!0.301\!\!\! & \!\!\!0.290\!\!\! & \!\!\!0.756\!\!\! & \!\!\!0.284\!\!\! & \!\!\!0.266\!\!\! & \!\!\!0.504\!\!\! \\
~~~+ Using Keyfacts & \!\!\!0.546\!\!\! & \!\!\!0.527\!\!\! & \!\!\!0.934\!\!\! & \!\!\!0.453\!\!\! & \!\!\!0.434\!\!\! & \!\!\!0.795\!\!\! \\\midrule
\textbf{\algname{}\,(GPT-4)}\!\! & \!\!\!\textbf{0.688}\!\!\! & \!\!\!\textbf{0.677}\!\!\! & \!\!\!\textbf{0.949}\!\!\! & \!\!\!\textbf{0.505}\!\!\! & \!\!\!\textbf{0.451}\!\!\! & \!\!\!\textbf{0.880}\!\!\! \\ \bottomrule
\end{tabular}
\end{center}
\vspace*{-0.45cm}
\caption{Fair comparison with G-Eval tuned for completeness and conciseness.} 
\label{table:geval-fair-comp}
\vspace*{-0.15cm}
\end{table*}

\vspace*{-0.1cm}
\section{Fair Comparison with G-Eval}
\label{sec:fair_g_eval}
\vspace*{-0.1cm}

For a truly fair comparison, we recognize the necessity to tune G-Eval for our two dimensions of completeness and conciseness. Hence, we opted to adjust G-Eval's prompts to align with the evaluation dimensions of FineSurE. We employed two variants of G-Eval tailored for the assessment of completeness and conciseness dimensions: one adjusted without the utilization of keyfacts, and another adjusted with the utilization of keyfacts. The two tuned prompts without using key facts are summarized in Figures \ref{fig:geval-completeness-wo-keyfacts} and \ref{fig:geval-conciseness-wo-keyfacts}, while those with using key facts are in Figures \ref{fig:geval-completeness-w-keyfacts} and \ref{fig:geval-conciseness-w-keyfacts}.

Table \ref{table:geval-fair-comp} compares FineSurE with two variants of G-Eval regarding completeness and conciseness. The findings illustrate that aligning evaluation criteria does not alter performance ("+Adjusted Criteria"), whereas the incorporation of human keyfacts notably enhances G-Eval ("+Using Keyfacts"). Nonetheless, FineSurE continues to outperform G-Eval, supporting that our fine-grained evaluation yields more precise assessments compared to the Likert-scale ratings used in G-Eval.

\begin{table*}[t]
\begin{center}
\footnotesize
\begin{tabular}{|L{5.3cm}  |X{3.6cm} |X{1.2cm} X{1.2cm} X{1.8cm}| }\toprule
Technique & Sentence-lev. & \multicolumn{2}{c}{Summary-lev.} & System-lev. \\
& bAcc (True Pos., True Neg.) & \!\!Pearson\!\! & \!\!\!\!Spearman \!\!\!\! & Rank\\ \midrule
Basic Prompt (Figure \ref{fig:faith_basic}) & 92.9\% (91.9\%, 94.0\%) &  0.856 & 0.850  & {0.933} \\ \midrule
~~~+ Inst. + Cat. (Figure \ref{fig:b_i_cat_faithfulness})              & {91.3\%} (90.1\%, 92.5\%) & 0.841 & 0.834 & {0.933} \\
~~~{{+ Inst. + Cat. + Rea.} (Figure \ref{fig:best_prompt_factuality})}  & {92.0}\% (91.9\%, 92.1\%) & {0.844} & {0.841} & 0.933 \\ 
~~~+ Inst. + Cat. + Evi. (Figure \ref{fig:b_i_c_e_faithfulness})       & 91.7\% (88.7\%, 94.7\%) & 0.829 & 0.821 & 0.933 \\ 
~~~+ Inst. + Cat. + Rea. + Evi. (Figure \ref{fig:b_i_c_r_e_faithfulness}) & 91.5\% (90.7\%, 92.3\%) & 0.839 & 0.836 & 0.933 \\ \bottomrule
\end{tabular}
\end{center}
\vspace*{-0.4cm}
\caption{Prompt engineering with Llama3-70B-Inst. for \textbf{faithfulness evaluation} using instruction (Inst), categorization (Cat), reasoning (Res), and evidence mapping (Evi) techniques.  }  
\label{table:ablation_faithfulness}
\vspace*{-0.0cm}
\end{table*}

\begin{table*}[t!]
\begin{center}
\footnotesize
\begin{tabular}{|L{4.0cm} |X{1.4cm} X{1.4cm} X{1.4cm} |X{1.4cm} X{1.4cm} X{1.4cm}|}\toprule
Dimension & \multicolumn{3}{c|}{(a) Completeness} & \multicolumn{3}{c|}{(b) Conciseness}  \\\midrule
Technique &  \multicolumn{2}{c}{Summary-level} & \!\!\!System-level\!\!\! & \multicolumn{2}{c}{Summary-level} & \!\!\!System-level\!\!\!  \\ 
& \!\!\!Pearson\!\!\! & \!\!\!Spearman\!\!\! & Rank & \!\!\!Pearson\!\!\! & \!\!\!Spearman\!\!\! & Rank \\ \midrule
Basic Prompt (Figure \ref{fig:b_other}) & 0.658 &  0.648 & 0.891 & 0.415 & 0.402 & {0.875} \\\midrule
~~~{+ Inst.} (Figure \ref{fig:best_prompt_keyfact}) & {0.775} & {0.747} & {0.881} & {0.445} & {0.444} & 0.786 \\
~~~+ Inst. + Rea  (Figure \ref{fig:b_i_r_other}) & 0.664 & 0.650 & 0.926 & 0.424 & 0.400 & 0.785 \\ \bottomrule
\end{tabular}
\end{center}
\vspace*{-0.4cm}
\caption{Prompt engineering with Llama3-70B-Inst. for \textbf{completeness and conciseness evaluation} using instruction (Inst) and reasoning (Res) techniques.  The values in parenthesis represent p-values.} 
\label{table:ablation_other}
\vspace*{-0.2cm}
\end{table*}

\section{Prompt Engineering}
\label{sec:prompt_engineering}

\begin{figure}[t!]
\begin{center}
\includegraphics[width=7.7cm]{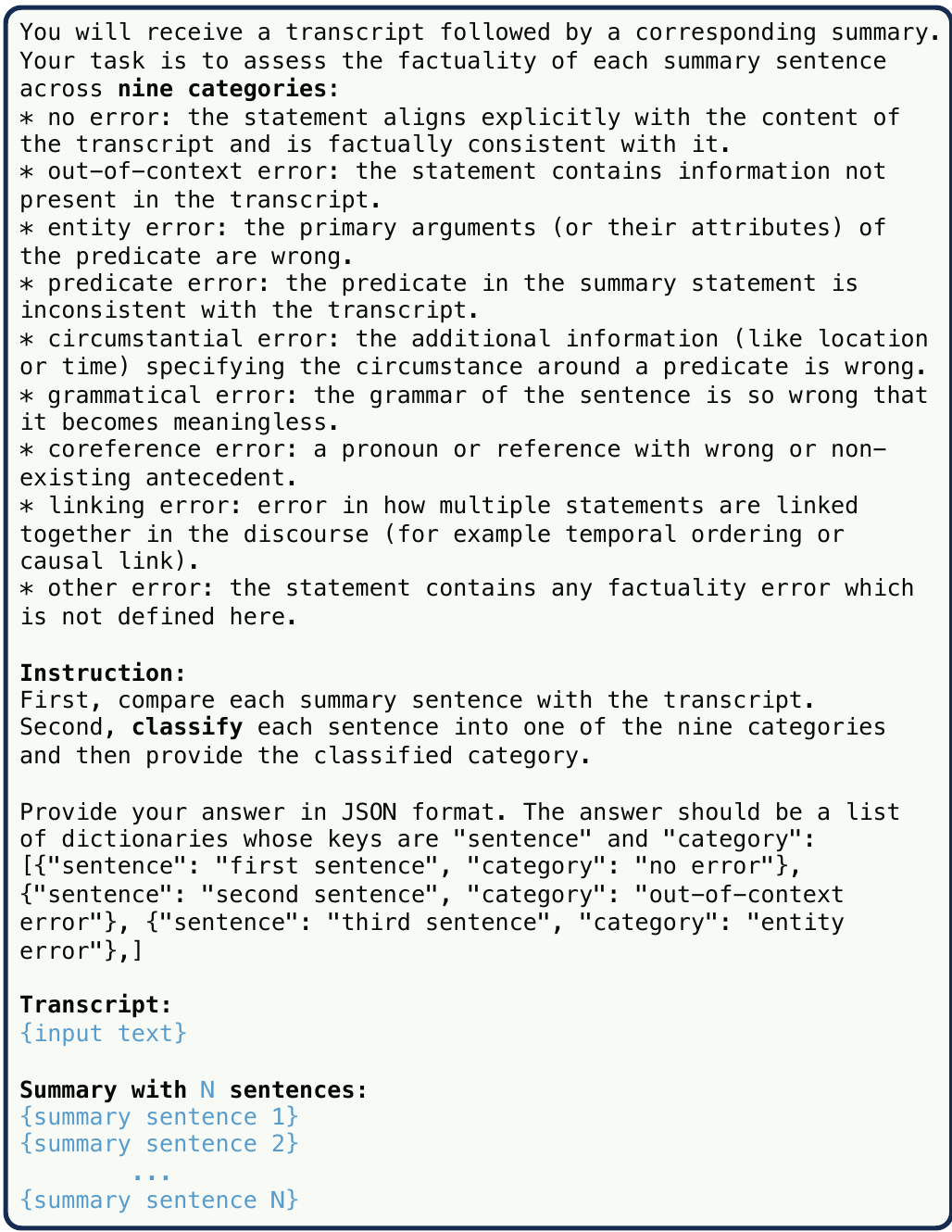}
\end{center}
\vspace*{-0.42cm}
\caption{Basic prompt with instruction format and categorization for faithfulness evaluation.}
\vspace*{-0.25cm}
\label{fig:b_i_cat_faithfulness}
\end{figure}

We test several prompt engineering techniques, including instruction format, solving categorization problem, reasoning, and evidence mapping. We summarize the agreement with human scores using different combination of prompting techniques with respect to faithfulness, completeness, and conciseness in Tables \ref{table:ablation_faithfulness} - \ref{table:ablation_other}, where the figure number \ref{fig:faith_basic}--\ref{fig:b_i_r_other} corresponding to each prompt is enclosed in parentheses in each row.

Regarding the prompt engineering for faithfulness, our most effective prompt involves the incorporation of three techniques: instruction format, solving categorization, and providing reasoning, as seen in in Figure \ref{fig:best_prompt_factuality}. This prompt shows the highest overall agreement with human judgments among those that included fine-grained evaluation, such as error categorization.

On the other hand, for completeness and conciseness, the inclusion of the reasoning step results in a decrease in agreement during evaluation. Additionally, categorization is unnecessary for keyfact alignment, given the presence of only two classes--"no matched" and "matched"--for each keyfact. Consequently, the most effective prompt involves solely utilizing the instruction format in Figure \ref{fig:best_prompt_keyfact}.

\begin{figure}[t!]
\begin{center}
\includegraphics[width=7.7cm]{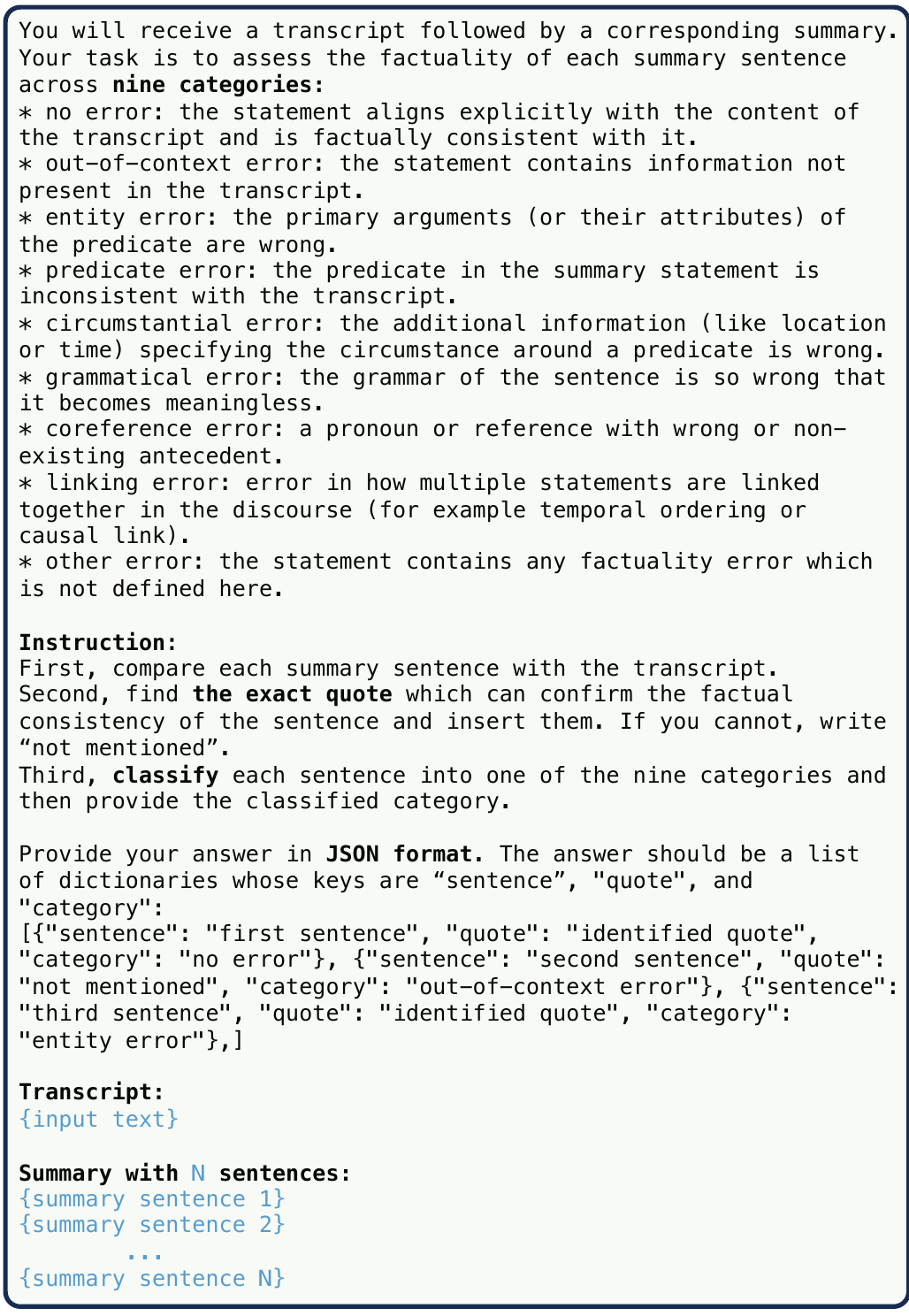}
\end{center}
\vspace*{-0.42cm}
\caption{Basic prompt with instruction format, categorization, and evidence mapping for faithfulness evaluation.}
\vspace*{-0.4cm}
\label{fig:b_i_c_e_faithfulness}
\end{figure}

\begin{figure}[t!]
\begin{center}
\includegraphics[width=7.7cm]{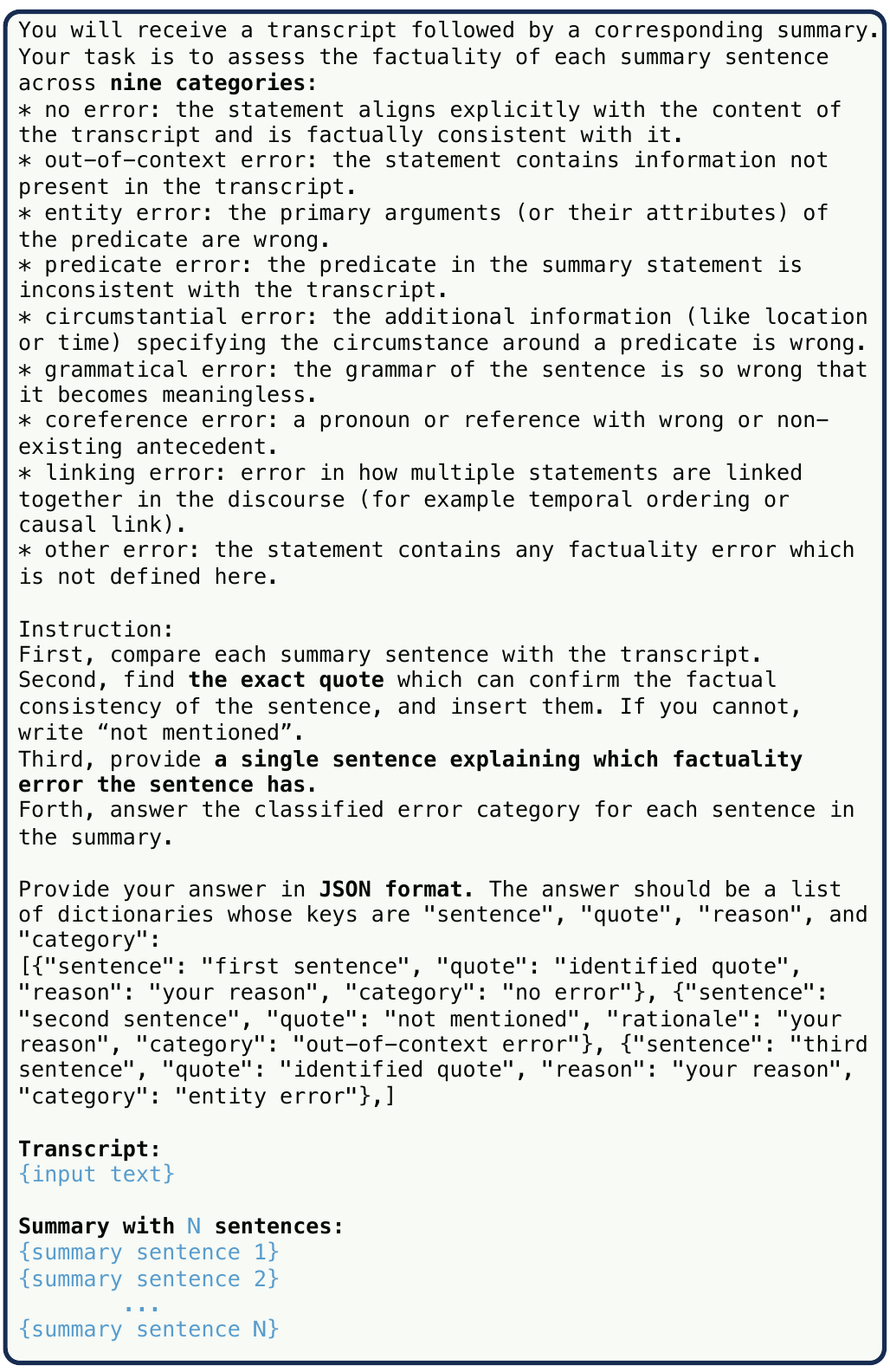}
\end{center}
\vspace*{-0.42cm}
\caption{Basic prompt with instruction format, categorization, reasoning and evidence mapping for faithfulness evaluation.}
\vspace*{-0.3cm}
\label{fig:b_i_c_r_e_faithfulness}
\end{figure}

\begin{figure}[t!]
\begin{center}
\includegraphics[width=7.7cm]{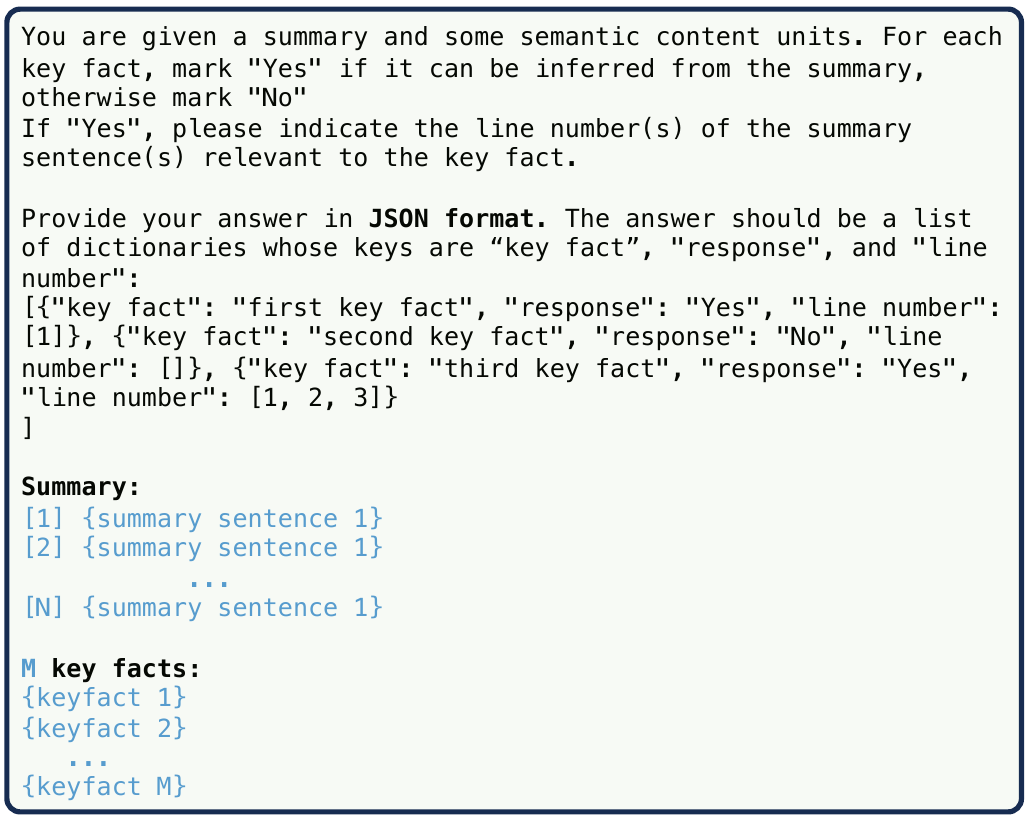}
\end{center}
\vspace*{-0.42cm}
\caption{Basic prompt for completeness and conciseness evaluation.}
\vspace*{-0.4cm}
\label{fig:b_other}
\end{figure}

\begin{figure}[t!]
\begin{center}
\includegraphics[width=7.7cm]{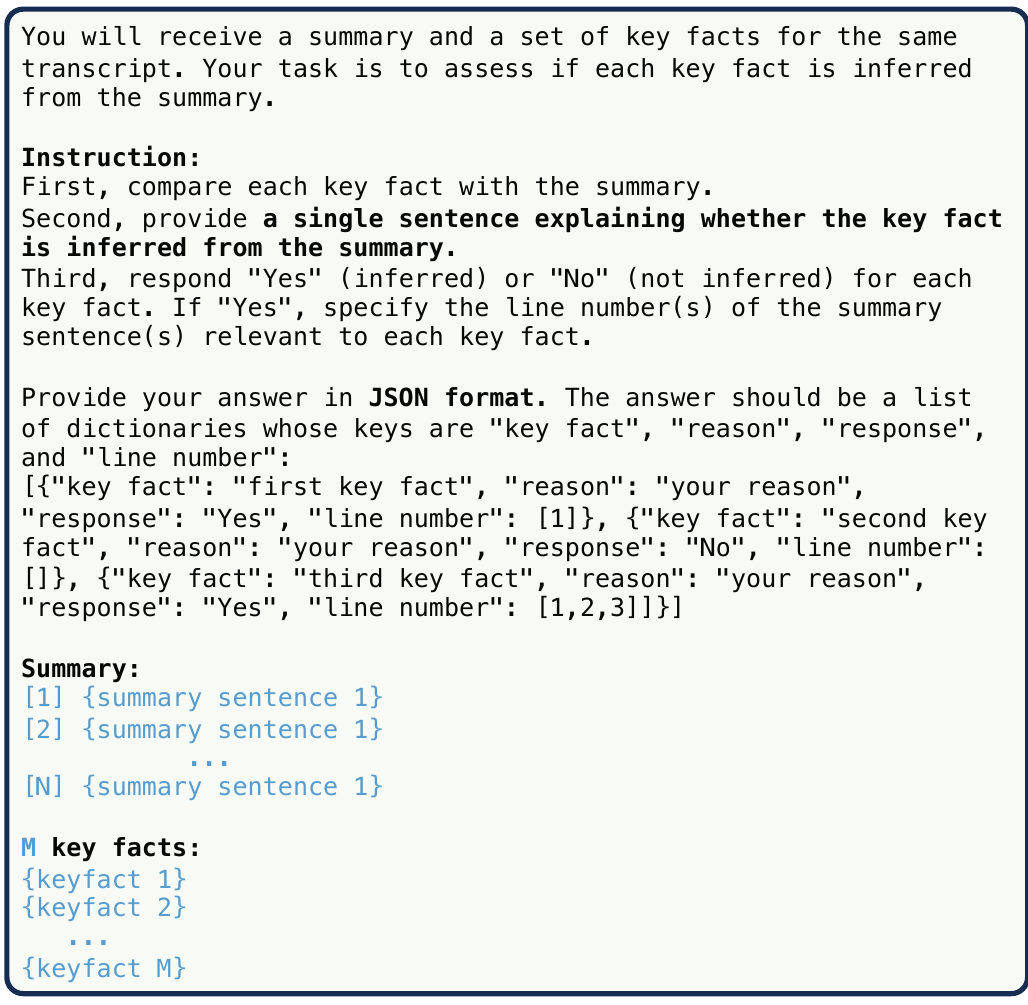}
\end{center}
\vspace*{-0.42cm}
\caption{Basic prompt with instruction format and reasoning for completeness and conciseness evaluation.}
\vspace*{-0.0cm}
\label{fig:b_i_r_other}
\end{figure}

\section{Summarization using LLMs}
\label{sec:llm_summary}

We consistently employ the prompt shown in Figure \ref{fig:prompt_summary} across all LLMs, generating model summaries based on the given input text. 

Furthermore, since the model summary is null when parsing fails, there is no hallucination, but also no summary sentences align with any key facts. Therefore, in cases where the LLMs produce incorrect JSON outputs that cannot be parsed, the scores for faithfulness, completeness, and conciseness are automatically set to 1.0, 0.0, and 0.0, respectively.

\begin{figure}[t!]
\begin{center}
\includegraphics[width=7.7cm]{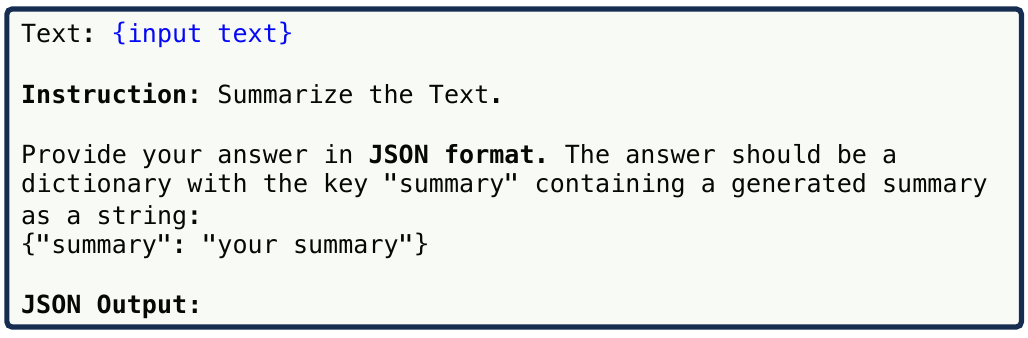}
\end{center}
\vspace*{-0.42cm}
\caption{Prompt to get the summary from LLMs.}
\vspace*{-0.2cm}
\label{fig:prompt_summary}
\end{figure}

\section{Extension of REALSumm}
\label{sec:extension_realsumm}

Although the REALSumm data contains human labels indicating which keyfacts are included in the model summary, there are no human labels indicating which summary sentences align with the keyfacts. The former is used to compute the ground-truth completeness score, while the latter is used to compute the conciseness score. Therefore, we conducted a human evaluation to verify which summary sentences align with the set of key facts. Specifically, three human annotators were asked to mark "yes" if at least one keyfact in the keyfact list could be inferred from each summary sentence, otherwise "no". This is quite simple task compared with faithfulness evaluation, since human ground-truth keyfacts are available in REALSumm data. The extended dataset is available with our FineSurE framework at \url{https://github.com/DISL-Lab/FineSurE}. 

\end{appendix}

\end{document}